\documentclass{article}
\usepackage{arxiv,times}
\usepackage[T1]{fontenc}
\usepackage[utf8]{inputenc}
\usepackage{amsmath,amssymb}
\usepackage{makecell}
\usepackage{graphicx,booktabs,array,tabularx}
\usepackage{xcolor,colortbl,microtype}
\usepackage{capt-of}
\usepackage{url,placeins}
\usepackage[colorlinks=true,citecolor=teal,linkcolor=teal,urlcolor=teal]{hyperref}
\newcommand{\projecturl}{\url{https://sirui-xu.github.io/InterMimicGen}}
\title{InterMimicGen: Scaling Humanoid\\Loco-Manipulation through\\Self-Evolving Motion Imitation}
\author{Yucheng Zhang$^{1\dag}$ \quad Sirui Xu$^{1\dag*}$ \quad Jinhong Li$^{1\dag}$ \quad Liuyu Bian$^{1\dag}$\\
\bfseries Anatulya Nandi$^{1}$ \quad Derek Zhang$^{1}$ \quad Xiangchen Liu$^{1}$ \quad Xueting Li$^{2}$\\
\bfseries Umar Iqbal$^{2}$ \quad Yu-Xiong Wang$^{1\ddag}$ \quad Liang-Yan Gui$^{1\ddag}$\\
\normalfont\small{$^{1}$University of Illinois Urbana-Champaign \quad $^{2}$NVIDIA}\\
\normalfont\small{$^{\dag}$Equal Contribution \quad $^{*}$Project Lead \quad $^{\ddag}$Equal Advising}\\
\normalfont\small\projecturl}
\newcommand{\ours}{InterMimicGen}
\usepackage{algorithm}
\usepackage{algpseudocode}
\newcommand{\D}{\mathcal{D}}
\usepackage{xcolor}
\definecolor{InterBlue}{HTML}{315A9D}
\definecolor{GenRose}{HTML}{B13B64}

\begin{document}
\maketitle
\vspace{-1cm}
\begin{center}
\includegraphics[width=\textwidth]{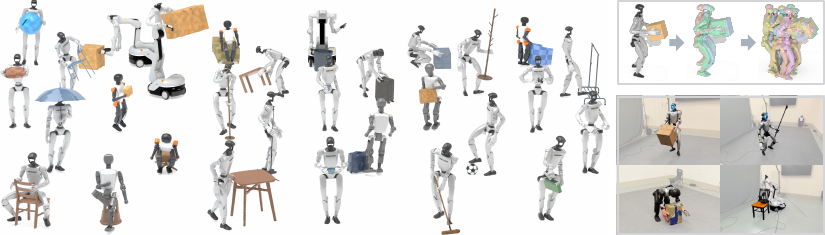}
\vspace{-0.7cm}
\captionof{figure}{\ours{} grows a finite set of human interaction captures into an expanding repertoire of executable robot motions. \textbf{Left:} One formulation turns the captures into executable motions across diverse objects, tasks, and robots, from humanoids to a wheeled mobile manipulator. \textbf{Top right:} Self-evolving motion imitation. A single motion grows round by round into many variants. \textbf{Bottom right:} The resulting motions run on real robots.}

\label{fig:teaser}
\end{center}
\vspace{-0.1cm}

\begin{abstract}
Captured human-object interactions provide rich supervision for humanoid loco-manipulation, but they are sparse, heterogeneous, and not directly executable by robots. We introduce \ours{}, a self-evolving motion-imitation framework in which robot motion data and a tracking policy improve each other. First, we consolidate motion-captured human-object interaction datasets and retarget them into humanoid robot references while preserving whole-body coordination and dexterous hand-object relationships. This produces a large and diverse humanoid robot reference collection for dexterous whole-body loco-manipulation. Second, we train a physics-based generalist tracker that executes these references in simulation on a humanoid with dexterous hands, covering a scale and diversity beyond prior humanoid tracking systems for loco-manipulation. Third, we close a data flywheel: each round makes small, task-preserving changes to where an interaction takes place and how the body performs it, fine-tunes the tracker on them, and keeps only the variants whose simulated execution completes the task, which seed the next round. With more iterations, these small edits compound into broader coverage around the sparse original demonstrations while preserving task semantics and motion quality. Experiments show contact-preserving retargeting across robot configurations, broad tracking with a single generalist policy, executable motions that keep growing over augmentation rounds, and transfer to real robots. \ours{} provides a unified path from heterogeneous human demonstrations to a continually expanding motion resource for humanoid robot learning.

\end{abstract}

\section{Introduction}
\label{sec:intro}
The value of human demonstrations for humanoid loco-manipulation extends beyond reproducing the trajectories they record. Each demonstration captures one coordinated solution in a high-dimensional interaction space, but even a large motion-capture corpus provides sparse samples of this space rather than dense coverage of their feasible variations. Covering these variations therefore requires either a policy that can infer the physically feasible manifold from these sparse samples, or a data-generation process that enriches the data with coordinated, physically valid variations.

A natural route is to train a tracking policy by collecting more references, but any fixed set of references still leaves these variations uncovered. When we look back, generalist trackers already generalize well in locomotion~\citep{sonic}, where a motion close to a demonstration usually remains feasible. Loco-manipulation is less forgiving. The robot must hold contact with an object while keeping its balance, so a small change in posture or object placement can break a grasp, push the hand into the object, or make the robot fall. As a result, a tracker that executes a demonstration may still fail on the same task where the trajectory is shifted by only a few centimeters (Sec.~\ref{sec:loop_exp}). Covering them therefore requires data that grows with the tracker rather than a fixed collection. We build this as a \emph{data flywheel}, in which executable data and the tracker grow together.

To instantiate this flywheel, we introduce \textbf{\ours{}}, a self-evolving motion-imitation data engine that couples human-object interaction retargeting, large-scale motion imitation, and iterative data expansion. We first consolidate heterogeneous human-object motion-capture data (Figure~\ref{fig:data_overview}) and retarget the interactions into humanoid robot references. Our retargeter keeps both long-horizon body-object relations and the finger contact that holds small objects: a whole-body solver shapes the grasp to follow the object's geometry rather than captured finger poses alone, and a fine-grained cleanup then refines the contact at the millimeter scale. The result is a large and diverse robot reference collection for dexterous whole-body loco-manipulation, as shown in Figure~\ref{fig:teaser}. Physics-based trackers then execute these references in simulation on a humanoid robot with dexterous hands, and the same trackers later grow together with their reference data through iterative data expansion.

The goal of this expansion is not to invent new tasks; each human interaction becomes a seed for physically distinct realizations of the same semantic behavior, in the spirit of \textcolor{GenRose}{MimicGen}'s demonstration-driven data expansion~\citep{mimicgen}. We vary object and body motion through small edits around previously validated interactions, such as placing a box farther left or crouching more deeply. Each round fine-tunes the tracker on the current references and edited candidates, executes the candidates in simulation, and retains only rollouts that complete without failure. Drawing on \textcolor{InterBlue}{InterMimic}'s insight that a policy can correct imperfect motion by learning from it~\citep{intermimic}, imitation repairs errors introduced by the edits. The successful rollouts then seed the next round. This feedback is the defining step of \textcolor{InterBlue}{Inter}\textcolor{InterBlue!50!GenRose}{Mimic}\textcolor{GenRose}{Gen}: corrected motions expand the reference set, and the expanded set enables the tracker to realize progressively larger variations.

Our contributions are threefold. First, we consolidate heterogeneous human-object interaction (HOI) data and retarget them into, to our knowledge, the most diverse humanoid robot reference collection (see Sec.~\ref{sec:eval}) for dexterous whole-body loco-manipulation. Second, using these references, we train a physics-based generalist tracker that realizes this repertoire on a target embodiment at a scale beyond prior humanoid tracking systems~\citep{hdmi,omniretarget,ultra,weave} for loco-manipulation. Third, we introduce a self-evolving augmentation-learning loop that expands this repertoire through coupled variations in object and whole-body motions, turning validated interactions into a growing resource for subsequent learning of whole-body controllers.

\section{Related Work}
\label{sec:related}
\noindent\textbf{Human-object interaction data.}
Human-object and human-scene capture datasets provide complementary interaction references across whole-body grasping, object tracking, bimanual manipulation, and scene-level coordination~\citep{grab,behave,intercap,himo,arctic,humoto,forcehoi,core4d,zhang2023neuraldome,zhao2023im,omomo}. These resources differ in object diversity, contact structure, and body and hand annotation, shaping the supervision available for imitation. Among them, InterAct~\citep{interact} consolidates heterogeneous interaction sources, while HiPHI~\citep{hiphi} adds extensive body-object capture. We combine their motion into a robot reference collection and expand physically executable coverage through retargeting and humanoid robot execution in simulation.

\noindent\textbf{Whole-body and dexterous motion retargeting.}
Motion retargeting adapts demonstrations across morphologies using learned mappings or optimization objectives built from skeleton structure, contact cues, and geometric correspondences~\citep{neuralretarget,skeletonretarget,contactretarget,sparseretarget,araujo2025gmr,umr}. For interaction, the relationships between body, object, hand, and scene must survive the change in embodiment rather than only the joint trajectory~\citep{avatarhoi,interactionretarget,omniretarget}. Dexterous teleoperation and synthesis study high-quality hand or hand-arm transfer~\citep{dexpilot,anyteleop,dexmachina,spider,dexplore}, but whole-body interaction also requires those contacts to remain consistent with locomotion and balance. Our retargeter therefore separates the problem into a coarse whole-body interaction solve and a local dense contact refinement, so hand motion is improved without disturbing the body-object motion.

\noindent\textbf{Physics-based imitation and whole-body manipulation.}
Physics-based imitation began with example-guided control of individual skills~\citep{deepmimic} and later scaled to shared controllers over diverse motion libraries~\citep{scalablecontrol,luo2023phc}. On physical humanoids, teleoperation and imitation enable whole-body control~\citep{omnih2o,twist}, while universal tracking policies and visuomotor extensions broaden the motion and sensing interfaces~\citep{unitracker,sonic,humanoidvla}. Object-centric extensions couple body and object dynamics for grasping and dynamic skills~\citep{physhoi,omnigrasp,skillmimic,maskedmanipulator,intermimic,interprior}, and then move to real-world humanoid loco-manipulation~\citep{hdmi,resmimic,demohlm,opt2skill,boxlocoman,ulc,dreamcontrol,visualmimic,physhsi,ultra}. Our work starts from reference tracking and extends it into a self-evolving data engine, with real deployment.

\noindent\textbf{Demonstration augmentation and iterative generation.}
A demonstration can be reused beyond its recorded trajectory~\citep{mimicgen}: collected segments preserve task intent, while execution under adapted conditions becomes valid training data. Later systems extend this execution-validated reuse to bimanual dexterity~\citep{dexmimicgen}, skill composition~\citep{skillmimicgen}, intervention-driven data collection~\citep{interventionalmimicgen,hilserl}, policy self-improvement~\citep{robocat}, and humanoid loco-manipulation~\citep{humanoidmimicgen}. A related line in character and humanoid control scales policies through generated experience or closed-loop motion generation~\citep{parc,dart,claims,hil}. Our setting differs in both domain and mechanism: we expand dexterous and agile humanoid loco-manipulation from motion-capture references, and we self-evolve through motion imitation itself instead of a separate kinematic motion synthesizer, which is often brittle for object interaction.

\section{Method}
\label{sec:method}

\label{sec:method_overview}
\begin{figure}[t]
    \centering
    \includegraphics[width=\linewidth]{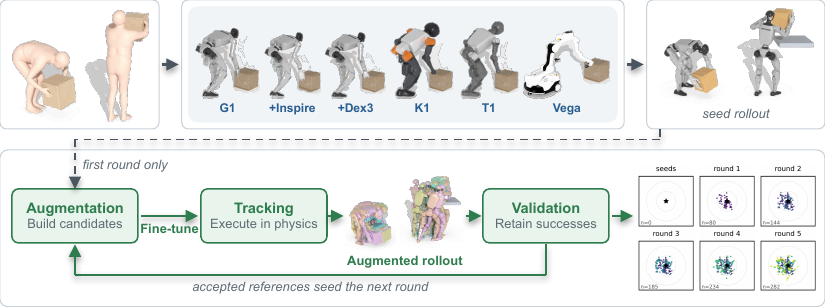}
    \vspace{-1em}
        \caption{\textbf{Top:} Reference construction. Human-object motion capture is retargeted into whole-body dexterous references, which a physics-based tracker imitates to produce seed rollouts. \textbf{Bottom:} Self-evolving motion imitation. Each round augments verified references into candidates, fine-tunes the tracker on them, executes the candidates in physics, and retains only successful rollouts as seeds for the next round; \textbf{Bottom right:} UMAP embeddings (Sec.~\ref{app:metrics_coverage}) of accepted augmentations expanding.}
    \label{fig:pipeline}
    \vspace{-1em}
\end{figure}

\ours{} turns captured human-object interactions into an expanding collection of robot references that a physics-based tracker can execute (Figure~\ref{fig:pipeline}). A reference $\boldsymbol{\xi}$ stores the target robot motion, object motion, and contact intent of one clip. We first consolidate heterogeneous captures (Sec.~\ref{sec:data}) and retarget them to robot embodiments (Sec.~\ref{sec:retarget}). A generalist tracker then realizes these references under robot dynamics (Sec.~\ref{sec:tracker}). Finally, self-evolving motion imitation edits how verified interactions are realized and keeps only the edits that the tracker executes in physics. Repeating this over rounds, with a tracker that keeps learning from the new motions, lets the verified references and the tracker grow together (Sec.~\ref{sec:augmentation}).

\subsection{A consolidated interaction reference collection}
\label{sec:data}
Because our learning loop expands skills already represented in the source motions, the seed collection should cover interaction semantics as broadly as possible. We combine InterAct~\citep{interact} and the object-interaction portion of HiPHI~\citep{hiphi} into a shared collection. InterAct contributes diverse tasks and object categories by consolidating public motion-capture datasets~\citep{grab,behave,intercap,zhang2023neuraldome,zhao2023im,omomo}, while HiPHI contributes longer synchronized human-object motions. Figure~\ref{fig:data_overview} summarizes this complementarity: the sources occupy different distributions of human motion, object geometry, and interaction space, and their union improves coverage on external motion probes. We convert both sources to a common body-object-hand representation and treat absent hand annotations as unknown rather than as evidence of no contact, and later refine the hand with retargeting.

We then keep only the interactions each target robot can physically support. Sequences that simulation cannot represent, such as bag hanging with an unmodeled strap, are removed, and platforms without dexterous hands exclude interactions whose success depends on articulated finger contact. The resulting references share one format for downstream training (Sec.~\ref{app:data}).

\subsection{Retargeting for dexterous whole-body interaction}
\label{sec:retarget}
Retargeting must satisfy constraints at two scales. The whole body must preserve support and body-object geometry over long horizons, while the hand must resolve millimeter-scale contact around small object regions. Optimizing both with one objective tends to loosen contact or over-constrain the body. We therefore solve the whole body first, with a finger-opposition term that shapes the grasp, and then remove local artifacts in a cleanup that preserves contact (Algorithm~\ref{alg:retarget}).

\noindent\textbf{Interaction-preserving whole-body retargeting.}
The coarse stage transfers body-object relationships through an interaction mesh~\citep{omniretarget} that connects sparse body landmarks with sampled object geometry, so matching the mesh preserves spatial interaction structure rather than only joint poses. It solves for the floating base and articulated body under the target robot's constraints, with sparse hand landmarks placing the wrists and palms. Because different motion clips might fail under different local constraints, we use several constrained solvers in turn and keep the first feasible solution. The solve keeps the source object trajectory fixed and includes the finger-opposition term below, so the hands close around the object while the body is solved.

\noindent\textbf{Finger opposition.}
Surface proximity alone can let several fingertips settle on the same side of an object, which does not hold it. We therefore add an opposition term to the whole-body solve. For fingertip $f$, let $\boldsymbol{n}_{t,f}$ be the normalized displacement from its selected object-surface point to the fingertip at time $t$, and let $\boldsymbol{n}_{t,\mathrm{thumb}}$ be the same direction for the thumb:
\begin{equation}
E_{\mathrm{cover}}(\boldsymbol{q}_t)
= \sum_{f\in\mathcal{F}_e} g_{t,f}
\big[ \boldsymbol{n}_{t,\mathrm{thumb}}^\top \boldsymbol{n}_{t,f}+m \big]_+^2,
\qquad [x]_+=\max(x,0),
\label{eq:coverage_loss}
\end{equation}
where $\mathcal F_e$ contains the supported thumb-finger pairs, $g_{t,f}$ activates a pair once the thumb and finger approach the object, and $m$ sets the opposition margin. The term penalizes pairs whose directions are not sufficiently opposed, so fingers  are promoted to approach from the side opposite the thumb.

\noindent\textbf{Geometry cleanup with contact preservation.}
The whole-body solve can still leave small penetrations, jitter, and contact drift. We therefore refine the robot and object trajectories with two successive Adam passes~\citep{Adam}. The first reduces robot-object, floor, and hand self-collision while regularizing deviation from the coarse motion, foot support, joint limits, and temporal smoothness. The second smooths the repaired trajectory under collision constraints and preserves hand-object relationships during contact. Each pass is retained only if its output satisfies deviation and quality checks relative to its input. Robots without articulated fingers omit finger-specific terms. Sec.~\ref{app:retargeting} gives the objectives, acceptance checks, and solver settings.

\subsection{A generalist for interaction motion imitation}
\label{sec:tracker}
The generalist tracker turns kinematic references into physically executable motions, and later serves as the validator of augmented references. It observes the robot state, upcoming reference frames, and object-relative geometry, including distance-decayed vectors from each body link to its nearest point on the object surface. It outputs joint targets for proportional-derivative (PD) control, and on the Inspire hands, coupled finger joints follow their driving joints through mimic-joint constraints instead of acting as independent actuators. We train a single policy $\pi_\theta$ with proximal policy optimization~\citep{ppo} across parallel environments, sharing parameters across motions and objects. The reward balances body motion, object motion, interaction geometry, and contact fidelity. Reference-state initialization exposes different phases of long sequences, and adaptive sampling emphasizes difficult motions. More details are presented in Sec.~\ref{app:tracking_details}.

\subsection{Self-evolving motion imitation}
\label{sec:augmentation}
A human capture shows what an interaction achieves but only one way of performing it. We therefore generate new ways of performing the same interaction. This follows the spirit of MimicGen~\citep{mimicgen}, which turns one demonstration into many training examples by replaying it under new conditions and keeping the successful replays, and here the replay is \textit{motion imitation} by our tracker. Each variant is derived from a verified parent reference $\boldsymbol{\xi}$ and must keep four properties of the parent, which form the invariant set $\mathcal I$: the interaction type, the intended contacts, the object, and the task outcome. Variants thus add new motions around each capture without changing its meaning.

\noindent\textbf{What an edit changes.}
An edit changes either where the interaction takes place or how the body performs it, as shown in Figures~\ref{fig:bio} and~\ref{fig:bio_more}. We describe it by coordinates $\boldsymbol{z}=(\boldsymbol{z}^{\mathrm{task}},\boldsymbol{z}^{\mathrm{body}})$ and construct a proposal $\tilde{\boldsymbol{\xi}}$ that keeps every invariant of its parent:
\begin{equation}
 \tilde{\boldsymbol{\xi}}=\operatorname{Augment}(\boldsymbol{\xi};\boldsymbol{z},\mathcal I),\qquad
 r_j(\tilde{\boldsymbol{\xi}},\boldsymbol{\xi})=0\quad\forall j\in\mathcal I ,
 \label{eq:interaction_operator}
\end{equation}
where $r_j$ measures how much the proposal violates invariant $j$. \textit{Object edits} $\boldsymbol{z}^{\mathrm{task}}$ move or turn the object's trajectory, for example placing a box further to the left, and leave the whole-body response to the tracker, which must reach, grasp, carry, and place the object accordingly. \textit{Body edits} $\boldsymbol{z}^{\mathrm{body}}$ keep the object path and change the posture that realizes it: a deeper crouch or a shifted and turned pelvis, a wider stance or a different toe angle, or a different elbow posture. In every case the hands stay on their original targets, so the contacts with the object are preserved. After each edit, we recompute the kinematics, velocities, and body-object interaction features of the new reference. Sec.~\ref{app:augmentation} defines the coordinates, their bounds, and the checks that each invariant of the parent still holds.

\noindent\textbf{Fine-tune, then keep only what executes.}
\label{sec:loop}
Each round starts from a set of verified parents. Let $\D_k$ be the verified references after round $k$, where $\D_0$ holds the seed rollouts of the base tracker on the retargeted references, and let $\mathcal P_k\subseteq\D_k$ be the parents of round $k$. The round edits the parents into proposals $\mathcal C_k$, fine-tunes the tracker $\pi_\theta$ on the verified references together with the proposals, since the proposals lie partly outside what it has learned, and then executes each proposal in physics and keeps the successful ones:
\begin{equation}
\begin{aligned}
\mathcal C_k
&= \operatorname{Augment}(\mathcal P_k),\\
\theta_{k+1}
&= \operatorname{Train}(\theta_k,\D_k\cup\mathcal C_k),\\
\mathcal B_k
&= \big\{\operatorname{Store}(\tilde{\boldsymbol{\xi}},\boldsymbol{\tau}):
\tilde{\boldsymbol{\xi}}\in\mathcal C_k,\
\boldsymbol{\tau}\sim\operatorname{Rollout}(\pi_{\theta_{k+1}},\tilde{\boldsymbol{\xi}}),\
V(\tilde{\boldsymbol{\xi}},\boldsymbol{\tau};\mathcal I)=1\big\},\\
\D_{k+1}
&= \D_k \cup \operatorname{Select}(\mathcal B_k).
\end{aligned}
\label{eq:archive}
\end{equation}
Here, $\operatorname{Train}$ continues the tracker's reinforcement learning on the proposed references. The validation function $V$ accepts a proposal only if its rollout reaches the end of the reference without early termination or a fall, passes motion-smoothness checks, and preserves the task outcome, such as holding or placing the object at the end (Sec.~\ref{app:loop_promotion}). Satisfying Eq.~\ref{eq:interaction_operator} alone does not make a proposal executable under robot dynamics. Fine-tuning expands the set of variations the tracker can execute: without it, only a small fraction of second-round proposals pass validation (Sec.~\ref{sec:ablation_finetune}). Thus, the loop improves the policy's capability as it expands the motion collection, rather than merely collecting variations the initial generalist could already track.

\noindent\textbf{Growing coverage over rounds.}
Accepted proposals become the parents of the next round, so the verified references and the tracker's skills grow together across more iterations. A proposal that fails is never edited again, so errors cannot accumulate across rounds. Each round picks a diverse set of the successful motions that reached farthest from their seed and moves each a fixed step further (Sec.~\ref{app:loop_recursion}). Coverage is measured independently of these edit coordinates (Sec.~\ref{app:body_range}).

\section{Experiments}
\label{sec:experiments}
We evaluate four questions. (\textbf{i})~Can one retargeter preserve both long-horizon body-object relations and fine finger contact, and does this yield cleaner references than prior retargeters (Sec.~\ref{sec:retarget_exp})? (\textbf{ii})~How broadly can trackers execute these references across objects and embodiments, and which interactions remain hard (Sec.~\ref{sec:tracking_exp})? (\textbf{iii})~Does growing the data together with the tracker expand what it can execute beyond the original references, and does this growth come from learning rather than from filtering more rollouts (Sec.~\ref{sec:loop_exp})? (\textbf{iv})~Do the resulting motions carry over from simulation to different real humanoid robots (Sec.~\ref{sec:hardware})?

\subsection{Experimental setup}
\label{sec:eval}
\begin{figure}
    \centering
    \includegraphics[width=1.0\linewidth]{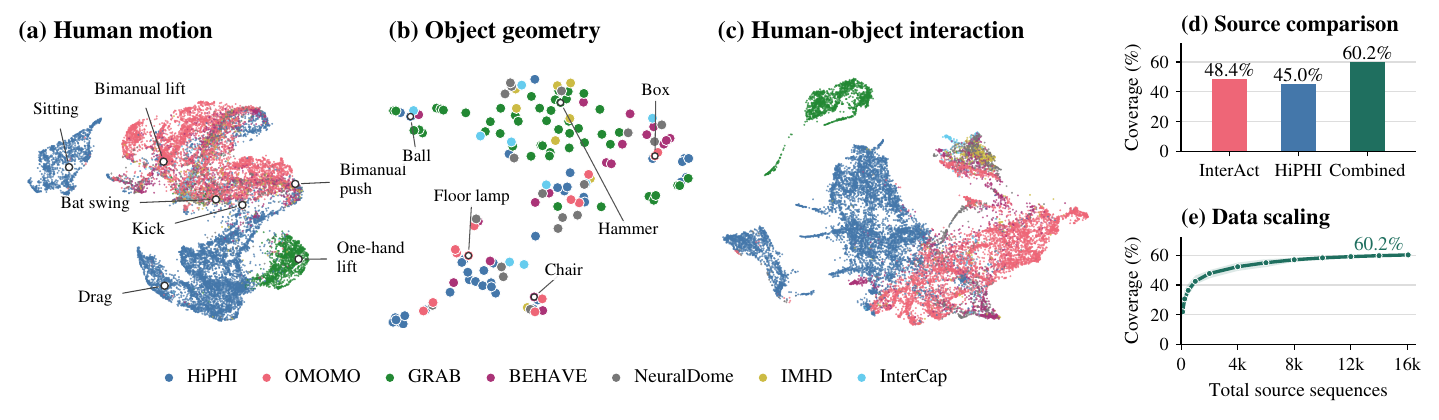}
    \vspace{-2em}
    \caption{\textbf{Our data has broad coverage.} (a-c) UMAP embeddings of human motion, object geometry, and human-object interaction of our collection, colored by source dataset. (d) How well our collection covers external motions from GRAIL~\citep{grail2026}, MOVIN~\citep{jang2023movin}, and ELMO~\citep{jang2024elmo}; a motion is covered when one of our sequences is close to it (Sec.~\ref{app:metrics_coverage}). (e) The same coverage as the number of our source sequences grows.}
    \label{fig:data_overview}
    \vspace{-1em}
\end{figure}

\noindent\textbf{Data and embodiments.}
We consolidate InterAct~\citep{interact} and HiPHI~\citep{hiphi} into a collection of 16,059 motions, 140.68 reference hours, and 157 object entries (Table~\ref{tab:corpus}), which covers a wide range of everyday interactions (Figure~\ref{fig:data_overview}). We retarget it to six robot configurations: G1~\citep{unitreeg1} with non-dexterous, Inspire~\citep{inspirehands}, or Dex3~\citep{unitreedex3} hands, Booster T1 and K1~\citep{boostert1,boosterk1}, and Dexmate Vega~\citep{dexmatevega}. Sec.~\ref{app:embodiments} details which experiments run on each embodiment.

\noindent\textbf{Metrics.}
Each experiment uses its own group of metrics, and Sec.~\ref{app:metrics} defines each of them once. Retargeted references are scored on plausibility and fidelity to the human motion: hand-contact preservation, in-hand slip, penetration, foot sliding, MPJPE, and joint acceleration. Tracking is scored by the success rate, the fraction of clips executed to the end without a fall or a large body, object, or interaction error, and by body and object errors on successful clips. For the augmentation loop, we report the growth of the verified reference set, the success of the tracker before and after self-evolution on the generated and original references, and the quality of their executions.

\noindent\textbf{Baselines.}
For dexterous retargeting, we compare with Weave~\citep{weave}, whose released retargeted references use the same embodiment with InterMimicGen. For retargeting without dexterous hands, we compare with OmniRetarget~\citep{omniretarget}, UMR~\citep{umr}, and ULTRA~\citep{ultra} on G1. For tracking, no publicly available tracker is trained on an interaction collection of comparable scale, so we compare our generalist with per-object specialists, each trained only on the references of one object, which serve as a practical upper bound. For the augmentation loop, we compare with the scene-scaling augmentation of ULTRA~\citep{ultra}, which also expands the training references of a humanoid tracker (Sec.~\ref{app:body_range}), and we isolate the effectiveness of augmentation with two controls: the base tracker before any augmentation and a frozen tracker that only filters new candidates without finetuning.

\begin{table}[t]
\caption{\textbf{Quantitative comparison on retargeting} on G1 with Inspire hands, on the 3,059 OMOMO clips. Ours shows better motion quality. Metrics are defined in Sec.~\ref{app:metrics_retarget}.}
\label{tab:retarget_vs_weave}
\centering
\footnotesize
\setlength{\tabcolsep}{0pt}
\begin{tabular*}{\linewidth}{@{\extracolsep{\fill}}lcccccccccccccc@{}}
\toprule
& \multicolumn{3}{c}{Hand contact} & \multicolumn{3}{c}{Penetration} & \multicolumn{3}{c}{Motion} & \multicolumn{3}{c}{Jnt.\ acc.\ (rad/s$^2$)$\downarrow$} & \multicolumn{2}{c}{Jnt.\ limit (\%)$\downarrow$} \\
\cmidrule(lr){2-4}\cmidrule(lr){5-7}\cmidrule(lr){8-10}\cmidrule(lr){11-13}\cmidrule(l){14-15}
& \shortstack{Frame\\(\%)$\uparrow$} & \shortstack{Hand\\(\%)$\uparrow$} & \shortstack{Slip\\(m/s)$\downarrow$}
& \shortstack{Obj.\\(\%)$\downarrow$} & \shortstack{Depth\\(cm)$\downarrow$} & \shortstack{Gnd.\\(\%)$\downarrow$}
& \shortstack{Foot sl.\\(\%)$\downarrow$} & \shortstack{MPJPE\\(cm)$\downarrow$} & \shortstack{EE\\(cm)$\downarrow$}
& All & Body & Fing. & Body & Fing. \\
\midrule
Weave & 96.1 & 97.1 & \textbf{0.225} & 67.4 & 2.62 & 31.9 & 62.0 & 9.54 & 12.6 & 12.2 & 6.3 & 19.3 & 0.6 & 33.4 \\
\rowcolor{gray!10}
Ours & \textbf{98.2} & \textbf{98.8} & 0.239 & \textbf{37.2} & \textbf{2.39} & \textbf{0.0} & \textbf{12.3} & \textbf{5.74} & \textbf{5.3} & \textbf{2.8} & \textbf{3.2} & \textbf{2.3} & \textbf{0.0} & \textbf{0.0} \\
\bottomrule
\end{tabular*}
\vspace{-1em}
\end{table}

\subsection{Retargeting quality and dexterous contact}
\label{sec:retarget_exp}
\noindent\textbf{Our retargeter produces cleaner dexterous references.}
Table~\ref{tab:retarget_vs_weave} compares our retargeting with Weave on G1 with Inspire hands. Our references keep hand contact slightly better, greatly reduce object penetration, foot sliding, and joint acceleration, and stay closer to the raw motion, at the cost of slightly higher in-hand slip. We attribute this to explicitly shaping the grasp and then additionally refining contact in a second cleanup, so better hands do not come at the cost of the body.

\noindent\textbf{Dexterous hands keep grasps that rigid hands can only approximate.}
Figures~\ref{fig:retarget_omni} and~\ref{fig:retarget_umr} compare against OmniRetarget~\citep{omniretarget} and UMR~\citep{umr} on shared source motions. With articulated fingers, our references preserve the grasp and hand-object relationship of the demonstration by explicit grasp modeling. When all methods use the same non-dexterous G1 (Sec.~\ref{app:rubber_hand}), each favors a different trade-off: ours is the smoothest and most accurate at the end effectors as the explicit grasping shape, OmniRetarget penetrates least, and UMR preserves the most contact.

\begin{figure}
    \centering
    \includegraphics[width=\linewidth]{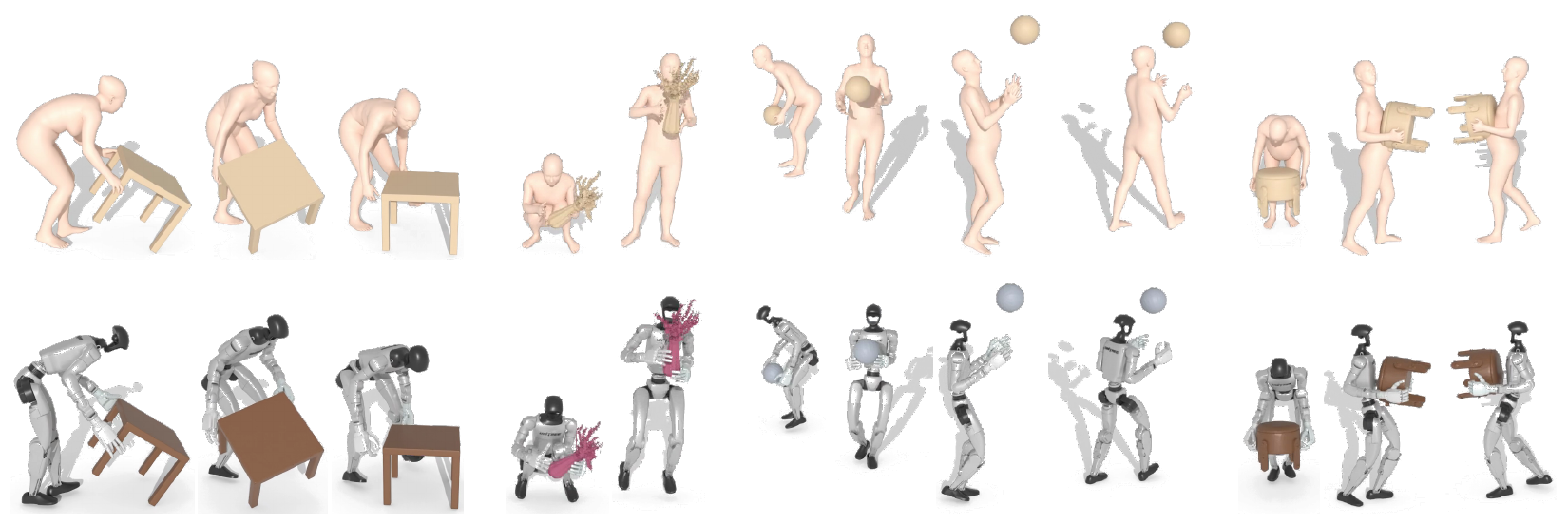}
    \vspace{-1em}
    \caption{\textbf{Qualitative evaluation on retargeting.} Captured human motion (top) and our G1 with Inspire hands reference (bottom), across interactions from different datasets.}
    \vspace{-1em}
    \label{fig:retarget_objects}
\end{figure}

\noindent\textbf{One formulation retargets across embodiments and objects.}
Figures \ref{fig:retarget_objects}, \ref{fig:retarget_more_objects}, \ref{fig:retarget_dexterous}, and~\ref{fig:retarget_embodiments} show interactions with objects of different shapes, a grasp retargeted to two dexterous hands, and one interaction retargeted to all six embodiments. The formulation is shared across robots: each configuration only declares its own landmarks and hand mappings, so a new robot needs only its configuration.

\begin{table}
\centering
\small
\setlength{\tabcolsep}{3pt}
\caption{\textbf{Quantitative results on motion tracking.} One generalist nearly matches the per-object specialists in body and object position error, while trailing them in success rate and object rotation. Metrics are defined in Sec.~\ref{app:metrics_tracking}; -- marks tasks the embodiment cannot perform.}

\label{tab:tracking}
\begin{tabular}{@{}lcccccccccccc@{}}
\toprule
& \multicolumn{4}{c}{Bimanual} & \multicolumn{4}{c}{Sitting} & \multicolumn{4}{c}{Grasping} \\
\cmidrule(lr){2-5}\cmidrule(lr){6-9}\cmidrule(l){10-13}
Embodiment & SR$\uparrow$ & MPJPE$\downarrow$ & Obj.$\downarrow$ & Rot.$\downarrow$ & SR$\uparrow$ & MPJPE$\downarrow$ & Obj.$\downarrow$ & Rot.$\downarrow$ & SR$\uparrow$ & MPJPE$\downarrow$ & Obj.$\downarrow$ & Rot.$\downarrow$ \\
\midrule
\multicolumn{13}{@{}l}{\textit{Specialists (one policy per object)}} \\
G1 + Inspire & 93.8 & 5.19 & 5.79 & 10.4 & 75.4 & 5.94 & 4.76 & 7.4 & 84.9 & 5.30 & 7.17 & 14.1 \\
G1 + Dex3  & 93.1 & 4.92 & 7.02 & 11.0 & 82.8 & 5.36 & 5.01 & 6.5 & 83.9 & 5.14 & 7.32 & 12.7 \\
G1   & 94.4 & 4.52 & 5.88 & 8.5  & 82.8 & 4.93 & 5.26 & 6.2 & -- & -- & -- & -- \\
K1   & 75.7 & 6.21 & 7.35 & 14.3 & 75.4 & 6.41 & 5.08 & 8.3 & -- & -- & -- & -- \\
\midrule
\multicolumn{13}{@{}l}{\textit{Generalist (one policy for all objects)}} \\
G1 + Inspire & 73.2 & 5.86 & 6.53 & 23.4 & 64.9 & 6.14 & 5.94 & 7.8 & 58.3 & 5.95 & 7.63 & 35.5 \\
\bottomrule
\end{tabular}

\end{table}

\subsection{Physics-based interaction tracking}
\label{sec:tracking_exp}
\begin{figure}
    \centering
    \includegraphics[width=\linewidth]{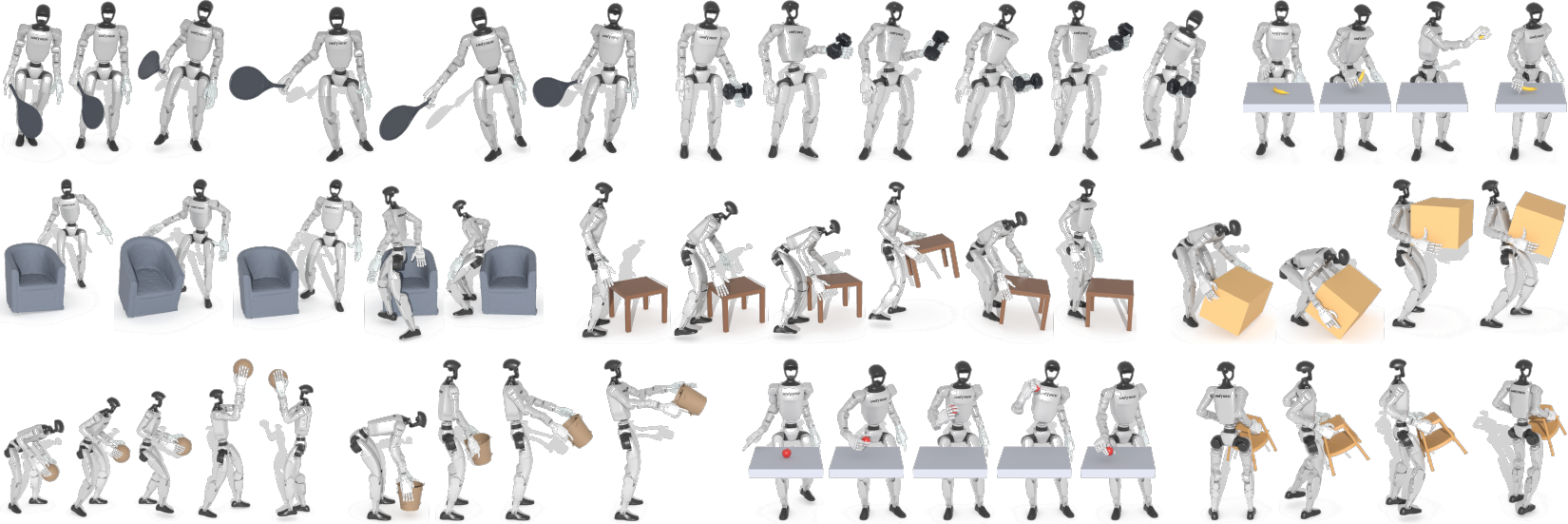}
    \vspace{-1em}
    \caption{\textbf{Physics-based tracking across objects.} Executed rollouts on G1 with Inspire hands for a diverse set of objects.}
    \label{fig:tracking_objects}
    \vspace{-1.5em}
\end{figure}

Table~\ref{tab:tracking} reports tracking on our constructed benchmark over 15 objects, grouped into bimanual, sitting, and grasping interactions. Bimanual interactions are tracked most reliably on every G1 configuration. Sitting and grasping are harder, since sitting shifts the body's weight onto the object and grasping requires finger contact on small object regions. The Booster K1 specialist trails the G1 specialists on bimanual interactions.
A single policy trained on the whole collection tracks all motions (Table~\ref{tab:tracking}, bottom). Its body error stays close to that of the specialists, whereas its success rate and object rotation error fall behind, most on grasping. Sharing one policy across objects costs mainly object control rather than body tracking, and fine grasps are where per-object capacity matters. Figures~\ref{fig:tracking_objects}, \ref{fig:tracking_embodiments}, and~\ref{fig:tracking_dexterous} show executed rollouts across objects, across embodiments, and with two dexterous hands.

\subsection{Does iterative generation expand learned competence?}
\label{sec:loop_exp}
We run five rounds of self-evolving motion imitation on bimanual and grasping interactions with G1 and Inspire hands, and on bimanual interactions with G1 and Dex3 hands, grouped from the successful rollouts in the tracking benchmark (Sec.~\ref{sec:tracking_exp}). Each round applies object edits, which move and turn where the object is placed relative to the seed's own execution, and body edits, which change the posture that realizes the interaction; later rounds edit only references verified in earlier rounds. We ask three questions: are the augmented references of good quality, does each round expand their range, and does training on them improve the tracker?
\begin{figure}
\centering
\vspace{-1em}
\includegraphics[width=\linewidth]{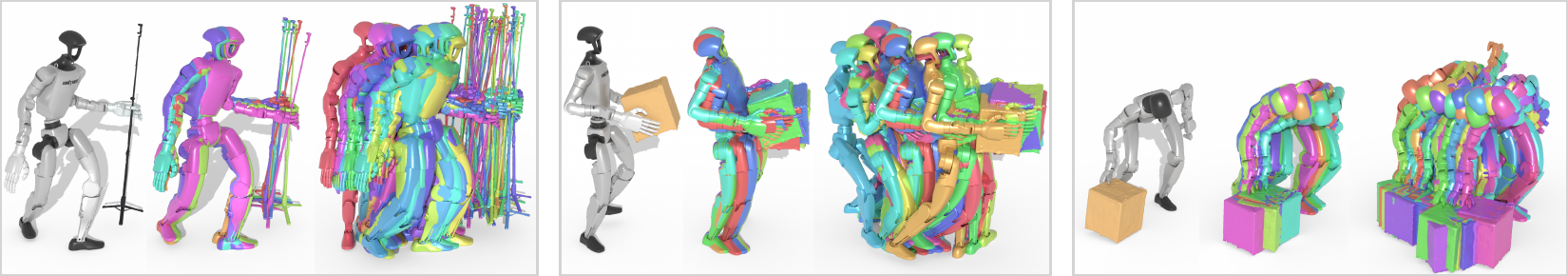}
\vspace{-1em}
\caption{\textbf{A single capture grows into many executable variants.} Each group shows the original reference followed by the verified augmentations accumulated by round 1 and by round 5.}
\label{fig:bio}

\end{figure}
\begin{table}
\caption{Over five rounds, the verified references grow \textbf{more than a hundredfold}, and the evolved tracker executes nearly all of them without losing the original ones. Growth: verified references relative to the original references. Success: rate at which the tracker before (Orig.) and after (Evol.) self-evolution. Quality: executions of original (Orig.) and augmented (Aug.) references.}
\label{tab:exec_augmented}
\centering
\footnotesize
\setlength{\tabcolsep}{0pt}
\begin{tabular*}{\linewidth}{@{\extracolsep{\fill}}lcccccccc@{}}
\toprule
& Growth (\texttimes) & \multicolumn{4}{c}{Success (\%)} & \multicolumn{3}{c}{Quality, Orig.\ $\to$ Aug.} \\
\cmidrule(lr){2-2}\cmidrule(lr){3-6}\cmidrule(l){7-9}
& & \multicolumn{2}{c}{Generated} & \multicolumn{2}{c}{Original} & Body acc. & Foot sl. & Hand jit. \\
\cmidrule(lr){3-4}\cmidrule(lr){5-6}
Setting & R1 $\to$ R5 & Orig. & Evol. & Orig. & Evol. & (rad/s$^2$)$\downarrow$ & (\%)$\downarrow$ & (mm)$\downarrow$ \\
\midrule
Inspire, bimanual & 21.8 $\to$ 150.5 & 52.1 & \textbf{98.4} & 100 & 100 & 11.4 $\to$ 12.4 & 1.18 $\to$ 1.87 & 1.00 $\to$ 1.17 \\
Dex3, bimanual & 21.3 $\to$ 142.0 & 59.3 & \textbf{98.7} & 100 & 100 & 13.9 $\to$ 16.0 & 1.14 $\to$ 2.21 & 1.12 $\to$ 1.30 \\
Inspire, grasping & 19.6 $\to$ 146.4 & 64.2 & \textbf{98.9} & 100 & 100 & 13.6 $\to$ 15.3 & 0.70 $\to$ 2.05 & 0.96 $\to$ 1.14 \\
\bottomrule
\end{tabular*}
\end{table}

\noindent\textbf{Augmented references stay close to the originals in quality.}
Averaged over all rounds, executions of augmented references stay close to those of the original references in body joint acceleration and hand jitter, while foot sliding rises from about 1\% to about 2\% of contact frames (quality columns of Table~\ref{tab:exec_augmented}). All three measures rise moderately in later rounds, which by design accept edits farther from the captures (Sec.~\ref{app:quality_rounds}).

\noindent\textbf{Each round expands the set of executable references.}
Every setting keeps accepting new spatial and body references: the verified reference set grows about \textbf{twentyfold} in the first round and more than a \textbf{hundredfold} by the fifth (growth column of Table~\ref{tab:exec_augmented}, Figures~\ref{fig:bio} and~\ref{fig:bio_more}; per-round values in Sec.~\ref{app:quality_rounds}). Body edits also widen the range along another axis, posture: their references spread over a region of body motion several times wider than the scene-scaling variants of ULTRA~\citep{ultra} (Sec.~\ref{app:body_range}). This expansion depends on learning. A frozen tracker that only filters candidates accepts almost none of the new candidates after the first round, because the new candidates start from first-round successes that already lie beyond its reach (Sec.~\ref{sec:ablation_finetune}).

\noindent\textbf{Self-evolution makes the tracker better.}
We replay every accepted reference and every original reference with the tracker before and after self-evolution, counting a reference as executed when at least one of three rollouts reaches its end (success columns of Table~\ref{tab:exec_augmented}). The original tracker fails on more than a third of the accepted references even though it is trained with large-scale data, whereas the evolved tracker executes nearly all of them, and both execute every original reference. Self-evolution therefore extends what the tracker can execute without losing what it could already do.

\subsection{Cross-embodiment execution and real-world transfer}
\label{sec:hardware}

\begin{figure}[tbp]
\centering
\includegraphics[width=\linewidth]{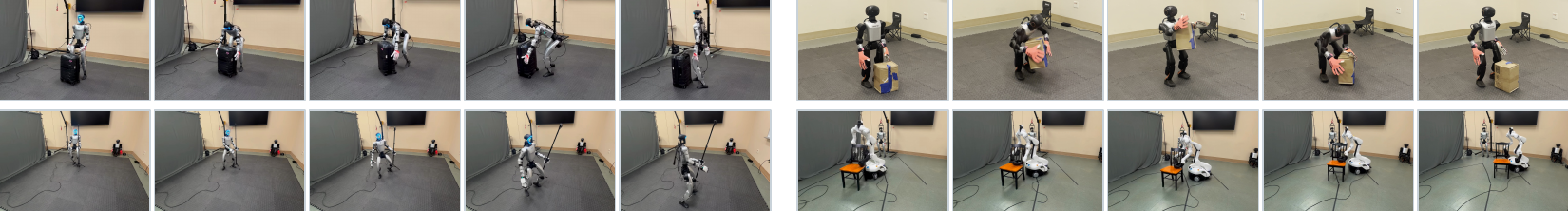}
\vspace{-1em}
\caption{\textbf{Real-world execution.} We show the motion from our paradigm transfer to different embodiments, from Unitree G1 pulling a suitcase and G1 with Inspire hands carrying a tripod; Booster K1 lifting and placing a box and Dexmate Vega relocating a chair.}
\label{fig:hardware}
\vspace{-1em}
\end{figure}

To verify that the trajectories produced by \ours{} work beyond simulation, we execute them on real robots. Each embodiment uses the parts of the framework it supports (Sec.~\ref{app:embodiments}), and each platform runs its own low-level controller, trained by us or publicly available (Sec.~\ref{app:hardware}). These controllers are blind to the object and track only the robot motion, yet our trajectories execute well on every platform, suggesting that they are physically consistent enough to transfer without object feedback. Figure~\ref{fig:hardware} shows box lifting and placement, suitcase pulling, tripod carrying, and chair relocation, with more videos on the \href{https://sirui-xu.github.io/InterMimicGen}{project page}.

\section{Discussion and Conclusion}
\label{sec:conclusion}
\ours{} treats a humanoid tracker as both a consumer and a producer of interaction data. A consolidated human-object motion collection supplies diverse task semantics, hand-aware retargeting transfers them to six robot configurations, and physics-based tracking grounds them in executable motion. The evolving loop then turns validated variations into new references. Every round enlarges the archive, and the fine-tuned tracker learns interactions beyond the base tracker's reach while keeping its seed skills. Learning is what keeps the loop going, since a frozen tracker stops accepting new variations after the first round.

\clearpage
{\small\noindent\textbf{Acknowledgments.} We thank Ziyin Wang for helping with the on-site deployment.\par}

\bibliography{references}
\bibliographystyle{arxiv}
\clearpage
\appendix
\renewcommand{\topfraction}{0.9}\renewcommand{\bottomfraction}{0.8}\renewcommand{\textfraction}{0.07}
\renewcommand{\floatpagefraction}{0.8}\setcounter{topnumber}{4}\setcounter{bottomnumber}{3}\setcounter{totalnumber}{6}
\makeatletter\setlength{\@fptop}{0pt}\setlength{\@fpsep}{12pt plus 4pt}\setlength{\@fpbot}{0pt plus 1fil}\makeatother
This appendix provides implementation details and extended evaluations that complement the main paper. Sec.~\ref{app:data} describes the data and embodiment scope, together with additional retargeting details in Sec.~\ref{app:retargeting}. Sec.~\ref{app:tracking_details} specifies the tracker's inputs and outputs, training setup, reference sampling, and rollout termination rules used to train and evaluate the generalist policy. Sec.~\ref{app:augmentation} details the augmentation protocol, including object edits, body edits, and how the tracker is fine-tuned over rounds. Sec.~\ref{app:metrics} defines the evaluation metrics. Sec.~\ref{app:exp} reports additional retargeting and augmentation experiments, and Sec.~\ref{app:hardware} documents the hardware protocol and details on real-world deployment.

\FloatBarrier
\section{Data and embodiments}
\label{app:data}
All embodiments draw on one consolidated collection of 16,059 sequences, 9,030 from InterAct and 7,029 from HiPHI. Every retargeted reference $\boldsymbol{\xi}$ provides the robot state, the object pose and twist, and the contact intent, so tracking and augmentation use the same format on every robot.

\begin{table}[!htbp]
\centering\small
\caption{\textbf{Reference collection} for G1 with Inspire hands. Object entries are dataset-object pairs, so an object that appears in two source datasets counts twice. Reference hours sum the frames of the selected robot references at 50\,fps and may include overlapping source segments.}
\label{tab:corpus}
\begin{tabular}{@{}lrrr@{}}
\toprule
Source & Object entries & Motions & Reference h\\
\midrule
InterAct~\citep{interact} & 117 & 9,030 & 17.82\\
HiPHI~\citep{hiphi} & 40 & 7,029 & 122.86\\
\midrule
Consolidated & 157 & 16,059 & 140.68\\
\bottomrule
\end{tabular}
\end{table}

\subsection{Embodiment scope}
\label{app:embodiments}
Table~\ref{tab:embodiment_scope} lists the studies run on each platform after the capability filtering of Sec.~\ref{sec:data}. Generalist training and iterative augmentation are evaluated mainly on G1 with dexterous hands, and the other platforms use the subsets they can perform. Motions a platform cannot support are excluded from its evaluation.

\begin{table}[!htbp]
\centering\small
\caption{Studies run on each platform. Configurations with dexterous hands use all data, including the dexterous benchmark; the generalist is trained on G1 with Inspire hands.}
\label{tab:embodiment_scope}
\setlength{\tabcolsep}{4pt}
\begin{tabular}{@{}lccccc l@{}}
\toprule
Platform & Retargeting & \makecell{Specialist\\tracking} & Generalist & \makecell{Augmentation\\loop} & \makecell{Sim-to-\\real} & \makecell[l]{Excluded\\tasks}\\
\midrule
G1 & \checkmark & \checkmark & -- & -- & \checkmark & dexterous\\
G1 + Inspire & \checkmark & \checkmark & \checkmark & \checkmark & \checkmark & none\\
G1 + Dex3 & \checkmark & \checkmark & -- & \checkmark & -- & none\\
Booster T1 & \checkmark & -- & -- & -- & -- & dexterous\\
Booster K1 & \checkmark & \checkmark & -- & -- & \checkmark & dexterous\\
Dexmate Vega & \checkmark & -- & -- & -- & \checkmark & dexterous, seated\\
\bottomrule
\end{tabular}
\end{table}

\FloatBarrier
\section{Retargeting}
\label{app:retargeting}
The retargeter of Sec.~\ref{sec:retarget} runs two stages (Algorithm~\ref{alg:retarget}): an interaction-preserving whole-body solve that includes finger opposition, followed by a two-pass geometry cleanup that preserves contact.

\noindent\textbf{Variables.}
The output is the robot trajectory $\boldsymbol q_{1:T}$, where $\boldsymbol q_t=(\boldsymbol q^{\mathrm{root}}_t,\boldsymbol q^{\mathrm{body}}_t,\boldsymbol q^{\mathrm{hand}}_t)$ holds the floating-base pose, the body joints, and the independent finger joints at frame $t$. The whole-body solve updates $\boldsymbol q_t$ frame by frame and keeps the source object trajectory $\boldsymbol x^o_{1:T}$ fixed. The cleanup updates the whole trajectory $\boldsymbol q_{1:T}$ together with small object-pose corrections $\Delta\boldsymbol x^o_{1:T}$.

\begin{algorithm}[t]
\caption{Retargeting one interaction clip to a target robot.}
\label{alg:retarget}
\small
\begin{algorithmic}[1]
\Require scaled human motion and object trajectory $\boldsymbol x^o_{1:T}$; robot configuration (landmarks, thumb-finger pairs $\mathcal F_e$, coupling maps)
\Ensure robot trajectory $\boldsymbol q_{1:T}$ and object trajectory $\boldsymbol x^o_{1:T}+\Delta\boldsymbol x^o_{1:T}$
\State \textit{Stage 1: interaction-preserving whole-body solve with finger opposition}
\For{$t=1,\dots,T$}
  \State update the opposition switches $g_{t,f}$ from finger-object distances and reference contact labels
  \State $\boldsymbol q_t\gets\arg\min E_{\mathrm{coarse}}$ (Eq.~\ref{eq:retarget_coarse}) under joint, support, and collision constraints, with $\boldsymbol x^o_t$ fixed and warm-started from $\boldsymbol q_{t-1}$
  \State \hspace{1em}try the solvers in turn and keep the first solution that passes the feasibility checks
\EndFor
\State \textit{Stage 2: geometry cleanup with contact preservation}
\For{pass $\in$ (penetration repair, collision-aware smoothing)}
  \Repeat
    \State Adam on $(\boldsymbol q_{1:T},\Delta\boldsymbol x^o_{1:T})$ with the pass objective (Table~\ref{tab:cleanup_weights}), then projection onto joint limits
    \State apply the acceptance check of the pass to the complete candidate against the pass input (Table~\ref{tab:cleanup_gates})
  \Until{the candidate passes, or all retry settings have been tried}
\EndFor
\State \Return $\boldsymbol q_{1:T}$, $\boldsymbol x^o_{1:T}+\Delta\boldsymbol x^o_{1:T}$
\end{algorithmic}
\end{algorithm}

\noindent\textbf{Interaction-preserving whole-body solve.}
The whole-body stage transfers body-object structure, rather than joint poses alone, from the scaled human reference to the robot. It builds an interaction mesh over body landmarks and sampled object geometry and minimizes
\begin{equation}
 E_{\mathrm{coarse}}=
 \lambda_I E_{\mathrm{interaction}}
 + \lambda_C E_{\mathrm{cover}}
 + E_{\mathrm{temporal}}
 + E_{\mathrm{posture}},
 \label{eq:retarget_coarse}
\end{equation}
subject to joint, support, and collision constraints, with the object trajectory fixed. The interaction term $E_{\mathrm{interaction}}$ matches the Laplacian coordinates of the robot-object mesh to the scaled source interaction~\citep{omniretarget}, while the temporal and posture terms regularize frame-to-frame motion and keep the posture plausible. We set $\lambda_I=10$, uniform over the mesh vertices. $E_{\mathrm{temporal}}$ penalizes the squared change of each coordinate of $\boldsymbol q_t$ from the previous frame's solution, with weight 2 for the floating base, legs, and waist and 1 for the arm and finger joints. $E_{\mathrm{posture}}$ penalizes the waist yaw and roll angles with weight 0.2 each. Sparse hand landmarks place the wrist and palm, and the opposition term $E_{\mathrm{cover}}$ of Eq.~\ref{eq:coverage_loss} shapes the fingers, with weight $\lambda_C=3$ and margin $m=1$. Because the cosine between two directions is at least $-1$, this margin keeps the term active for every active pair, so the term pulls each pair toward full opposition.

\noindent\textbf{When opposition is active.}
The switch $g_{t,f}$ turns the opposition term on only when a grasp is forming. It activates when the thumb and at least one other finger are within 10\,cm of the object, holds for 15 frames, and resets only after the fingers move more than 15\,cm away. Contact labels in the reference can also trigger it.

\noindent\textbf{Solvers.}
Different clips fail under different local constraints and solvers, so we solve the same interaction-mesh objective with several constrained solvers, tried in turn: MOSEK, Clarabel, ECOS, CVXOPT, OSQP, and SCS, plus box-bound variants of Clarabel, OSQP, and SCS that replace the local Euclidean step bound with componentwise bounds. The coarse stage is solved frame by frame, warm-started from the previous frame. The first solution that passes the feasibility checks initializes the cleanup stage.

\noindent\textbf{Surface correspondences.}
The opposition term needs an object-surface point for every active fingertip. For fingertip landmark $\boldsymbol{p}_{t,f}$ and selected object-surface point $\boldsymbol{s}_{t,f}$ on the object's collision geometry, we define $\boldsymbol{v}_{t,f}=\boldsymbol{p}_{t,f}-\boldsymbol{s}_{t,f}$ and the direction $\boldsymbol{n}_{t,f}=\boldsymbol{v}_{t,f}/\|\boldsymbol{v}_{t,f}\|$ of Eq.~\ref{eq:coverage_loss}. The term depends on the robot configuration through the wrist and hand forward kinematics and is purely geometric: it compares the thumb direction with each active opposing fingertip direction, which pushes the fingers to contact the object from complementary sides.

\noindent\textbf{Actuation mappings.}
One hand formulation serves different robots because each configuration declares its own kinematics to the shared objective. A dependent joint, such as a coupled finger joint of the Inspire hands~\citep{inspirehands}, is written as $q_j=\alpha_j q_{i(j)}+\beta_j$ and reconstructed inside the differentiable hand model before forward kinematics. Its derivative is folded into the Jacobian of the independent joint, $J_{i(j)} \leftarrow J_{i(j)} + \alpha_j J_j$, for the landmark, collision, and opposition terms. Each configuration also declares its body and fingertip landmarks and its supported thumb-finger pairs $\mathcal F_e$. Hands with fewer fingers~\cite{unitreedex3} use the pairs they have.

\noindent\textbf{Geometry cleanup with contact preservation.}
The whole-body solution can still leave millimeter-scale penetration and jitter. The cleanup optimizes the robot joints, the floating base, and small object-pose corrections with two successive Adam passes, each followed by projection onto joint limits. Both passes are strongly regularized toward their input. Table~\ref{tab:cleanup_weights} lists the default weights. Penetration repair runs 300 Adam and 60 projection steps. It penalizes robot-object penetration with a 2\,mm clearance margin, robot-floor and object-floor penetration, and hand self-collision beyond the overlap already present in its input, while fidelity and smoothness terms keep the edit small and free of new roughness. Collision-aware smoothing then runs 400 Adam and 60 projection steps. It penalizes acceleration and jerk of the body, hands, root, object, and body landmarks, keeps the collision penalties so that smoothing cannot reintroduce penetration, and weakens data fidelity to 0.02 at frames with sudden hand jumps so that these jumps can be smoothed away.

\noindent\textbf{How contact is preserved.}
Two terms protect contact. In both passes, a contact-keeping term penalizes robot points that were near contact in the input and move more than 2\,cm away from the object. Smoothing also keeps the hand-base pose fixed relative to the object at frames in contact, detected as an object distance below 1.5\,cm, with a relative-pose scale of 2\,cm. These terms preserve and stabilize existing contact but do not attract fingertips to new surface points. Grasp shape comes from the opposition term of the whole-body solve.

\begin{table}[t]
\centering
\small
\begin{minipage}[t]{0.52\linewidth}
\caption{\textbf{Cleanup objective weights} (default settings). ``acc/jerk'': the weight applies to both. \checkmark: term used, weight unlisted.}
\label{tab:cleanup_weights}
\end{minipage}\hfill
\begin{minipage}[t]{0.45\linewidth}
\caption{\textbf{Cleanup acceptance checks.} Repair is measured against the whole-body solution, smoothing against the repaired trajectory.}
\label{tab:cleanup_gates}
\end{minipage}
\begin{minipage}[t]{0.52\linewidth}
\centering
\setlength{\tabcolsep}{2.5pt}
\begin{tabular}[t]{@{}lcc@{}}
\toprule
Term & Repair & Smoothing \\
\midrule
\multicolumn{3}{@{}l}{\textit{Collision}} \\
Robot-object penetration & 3000 & 3000 (new) \\
Robot-floor penetration & 3000 & 3000 (new) \\
Object-floor penetration & 5000 & 5000 \\
Hand self-collision (new) & 300 & 300 \\
\midrule
\multicolumn{3}{@{}l}{\textit{Fidelity and contact}} \\
Trajectory deviation & \checkmark & 1 \\
Body-landmark drift & 2 & 2 \\
Planted-foot drift & 30 & 10 \\
Joint limits & 1 & 1 \\
Contact keeping & \checkmark & 20 \\
Hand pose rel.\ object & -- & 0.5 \\
\midrule
\multicolumn{3}{@{}l}{\textit{Smoothness}} \\
Body joints acc/jerk & 20 & 40 \\
Hand joints acc/jerk & 20 & 80 \\
Root translation / rotation & 4000 (acc) & 4000 / 300 \\
Object translation / rotation & 8000 (acc) & 16000 / 300 \\
Body landmarks & 2000 (acc) & 1000 \\
Edit roughness & 800 & -- \\
\bottomrule
\end{tabular}
\end{minipage}\hfill
\begin{minipage}[t]{0.45\linewidth}
\centering
\setlength{\tabcolsep}{2.5pt}
\begin{tabular}[t]{@{}lcc@{}}
\toprule
Maximum change & Repair & Smoothing \\
\midrule
Root position & 5\,cm & 5\,cm \\
Body joint & 0.5\,rad & 0.5\,rad$^{a}$ \\
Hand joint & 1.5\,rad & 1.5\,rad \\
Object position & 6\,cm & 4\,cm \\
Object rotation & 15$^\circ$ & 25$^\circ$ \\
Non-hand body position & 15\,cm & 12\,cm \\
Hand body / root endpoint & -- & 20 / 2\,cm \\
Object penetration & +3\,mm & +2\,mm \\
Non-foot floor penetration & +3\,mm$^{b}$ & +2\,mm \\
Jerk metrics & 3$\times$ & 1.5$\times$ \\
Fingertip-aperture acc. & -- & +0.02 \\
\midrule
Retained contact frames & $\geq$50\% & $\geq$60\% \\
\bottomrule
\end{tabular}\\[2pt]
{\footnotesize $^{a}$RMS $\leq$ 0.08\,rad. $^{b}$Above $\max(\text{input},3\,\text{mm})$.}
\end{minipage}
\end{table}

\noindent\textbf{Acceptance checks.}
Each pass accepts a candidate only after the complete trajectory passes its check against that pass's input: the repair check compares with the whole-body solution, and the smoothing check compares with the repaired trajectory (Table~\ref{tab:cleanup_gates}). The checks bound how far the root, joints, object, and body may move, how much penetration may grow, and how many contact frames must survive, which applies when the input has at least five. Repair must also improve penetration: if the input's maximum penetration exceeds 3\,mm, the output must reach at most 3\,mm or cut that maximum by at least 25\%. Smoothing must not worsen the aggregate jerk score. Jerk limits allow each metric to grow by the listed factor plus a small fixed margin. Both checks reject nonfinite results, quaternion-norm errors of at least $10^{-6}$, new joint-limit violations, and objects that sink more than 5\,mm below their starting height or the floor.

\noindent\textbf{Retargeting to wheeled robots.} For wheeled embodiments such as the Dexmate Vega, we keep the interaction-mesh objective and adapt the configuration space and landmark correspondences. The planar base translation and yaw are optimized jointly with the upper body, while the chassis height stays at its nominal value and the base roll and pitch are fixed at zero. Human leg and foot landmarks are removed from the interaction mesh and the foot-contact constraints are disabled, so the base, torso, and arms together preserve the demonstrated manipulation without passing changes in human pelvis height to the chassis. The three torso joints are constrained only through the mapped shoulder, elbow, wrist, and hand landmarks rather than by matching human spine angles. Temporal regularization and bounds on frame-to-frame changes of the upper-body joints and the planar base limit abrupt motion, and the wheel joints stay fixed during this kinematic stage. We discuss how the trajectories can be used by the real controller in Sec.~\ref{app:hardware}.

\FloatBarrier
\section{Generalist interaction tracker}
\label{app:tracking_details}
The tracker $\pi_\theta$ of Sec.~\ref{sec:tracker} is a privileged tracker, trained with PPO on G1 with Inspire hands using the settings in Table~\ref{tab:app_tracker_configuration}. Its observation, action, and reward define each control step, and its training and evaluation protocol decides which references it sees and when a rollout ends.

\subsection{Observation, action, and reward}
\label{app:tracker_policy}
\noindent\textbf{Observation.}
The policy observes robot proprioception, heading-local differences to upcoming reference frames, object-relative state, and body-object geometry. The geometric part encodes, for each body link $i$, the heading-local displacement $v_i$ to its nearest object-surface point as a distance-decayed vector:
\begin{equation}
 \psi(v_i)=\frac{v_i}{\|v_i\|_2+\epsilon}
 \exp(-\kappa\|v_i\|_2),
 \qquad \epsilon=10^{-6},\quad \kappa=5\,\mathrm{m}^{-1}.
 \label{eq:app_geometry_observation}
\end{equation}
The feature is computed for both the current and the reference geometry. It makes the policy sensitive to nearby interaction structure and applies unchanged to hands with different joint spaces.

\noindent\textbf{Action and PD control.}
The policy outputs joint-position offsets $a_t$, which a PD controller converts to torques:
\begin{equation}
 \tau_t=\operatorname{clip}_{\tau_{\max}}
 \left[K_p\big(q_{\mathrm{rest}}+s_a a_t-q_t\big)-K_d\dot q_t\right],
 \label{eq:app_pd}
\end{equation}
where $q_t$ and $\dot q_t$ are the joint positions and velocities, $q_{\mathrm{rest}}$ is the rest pose, and the gains $K_p$ and $K_d$, action scale $s_a$, and torque limit $\tau_{\max}$ are embodiment-specific. On the Inspire hands, coupled finger joints follow their driving joints through the mimic-joint mapping, which is reconstructed before simulation, instead of acting as independent actuators.

\noindent\textbf{Reward.}
The reward multiplies body, object, interaction, and contact tracking terms, so success in one term cannot offset failure in another, and adds regularization:
\begin{equation}
 r_t^{\mathrm{track}}=s_r
 r_t^{\mathrm{body}}r_t^{\mathrm{object}}
 r_t^{\mathrm{interaction}}r_t^{\mathrm{contact}},
 \qquad
 r_t=r_t^{\mathrm{track}}+\lambda_t r_t^{\mathrm{regularization}},
 \label{eq:app_reward}
\end{equation}
where $s_r$ scales the tracking reward and $\lambda_t$ weights the regularization. The body and object terms track pose and velocity, and the interaction and contact terms maintain the intended body-object relation. The interaction error compares body-to-surface vectors directly and weights close pairs most:
\begin{equation}
 w_{ij}=\frac{\max(\|d_{ij}\|_2^2,\eta)^{-1}}
 {\sum_{a,b}\max(\|d_{ab}\|_2^2,\eta)^{-1}},
 \qquad
 E^{\mathrm{track}}_{\mathrm{int}}=
 \frac12\sum_{i,j}(w_{ij}+\hat w_{ij})
 \|d_{ij}-\hat d_{ij}\|_2^2,
 \label{eq:app_interaction_error}
\end{equation}
where $d_{ij}$ and $\hat d_{ij}$ are the current and reference body-surface vectors, $\hat w_{ij}$ is the same weight computed from $\hat d_{ij}$, and $\eta$ floors the squared distance. The regularization includes an object-energy factor that penalizes the simulated linear and angular acceleration of the object, obtained by differencing velocities at the control interval.

\subsection{Training and evaluation protocol}
\label{app:verified_protocol}
\noindent\textbf{Object-compatible adaptive sampling.}
Each simulated environment holds one object asset, so it samples only the references compatible with that object. Within this compatible set $\mathcal R_o$, reference $i$ is drawn with probability
\begin{equation}
 p(i\mid o)=\frac{\rho}{|\mathcal R_o|}
 +(1-\rho)\frac{[\operatorname{clip}(\bar c_i,c_{\min},1)]^{-\beta}}
 {\sum_{j\in\mathcal R_o}[\operatorname{clip}(\bar c_j,c_{\min},1)]^{-\beta}},
 \quad \rho=0.25,\; c_{\min}=0.05,\;\beta=0.5,
 \label{eq:app_motion_sampling}
\end{equation}
where $\bar c_i$ is the moving-average completion of reference $i$. The uniform share $\rho$ keeps every reference in training, and the second term shifts the remaining mass toward clips the policy rarely completes.

\noindent\textbf{Start frames and termination.}
Reference-state initialization starts training episodes at sampled frames before the terminal hold, exposing different phases of long sequences. A rollout terminates on a fall, a non-hand body-position error above 0.5\,m, an object-position error above 0.5\,m, a normalized interaction error above one, or a nonfinite state. Physical tests use a short initialization grace interval, and terminal-hold frames are excluded from source-duration and coverage accounting.

\noindent\textbf{Contact-prioritized hand reward.}
The generalist scores grasp contact during reference-active hand phases. For distal pad $f$, let $\boldsymbol b_f=\boldsymbol p_f-\boldsymbol u_f$ point from its nearest object-surface sample to the pad center, and let $\Delta_f=\max(\|\boldsymbol b_f\|_2-r_f,0)$ be the gap beyond its collision-mesh-derived effective radius $r_f$. We compute proximity $P_f=\exp[-(\Delta_f/\sigma)^2]$, contact proxy $C_f=S_f[1-\Delta_f/d_c]_0^1$, and thumb-finger opposition $O_f=[(1-\hat{\boldsymbol b}_h^\top\hat{\boldsymbol b}_f)/2]_0^1$, where $h$ denotes the thumb, $\sigma=0.05\,\mathrm{m}$, and $d_c=0.025\,\mathrm{m}$. Here $S_f$ thresholds the sum of absolute simulator contact-force components on the distal link at $0.1$; proximity filtering reduces credit for contact away from the target object but does not identify the contacting body. For non-thumb fingers $f$, define
\begin{equation}
 A=\max_f\sqrt{P_hP_f},\quad E=\max_f(\sqrt{P_hP_f}\,O_f),\quad
 C_{\mathrm{opp}}=C_h\max_f(C_fO_f),\quad C_{\mathrm{any}}=\max_{f\in\mathrm{pads}}C_f.
\end{equation}
The palm proxy $C_{\mathrm{palm}}=S_{\mathrm{palm}}[1-\Delta_{\mathrm{palm}}/d_c]_0^1$ uses the same proximity filter, where $\Delta_{\mathrm{palm}}$ is the palm-to-object-surface gap. The per-hand grasp score is
\begin{equation}
 G=0.05 A+0.10 E+0.65 C_{\mathrm{opp}}
       +0.15 C_{\mathrm{any}}+0.05 C_{\mathrm{palm}}.
 \label{eq:app_hand_score}
\end{equation}
The active-hand scores are combined with a limited support-contact fallback, then converted to a bounded reward factor with strength $0.25$. This factor replaces the earlier hand-tracking factor rather than multiplying on top of it. The score favors the same thumb-finger opposition encouraged by $E_{\mathrm{cover}}$ in Eq.~\ref{eq:coverage_loss}.

\begin{table}[!htbp]
\centering\small
\caption{Generalist tracker settings.}
\label{tab:app_tracker_configuration}
\begin{tabularx}{\linewidth}{@{}p{0.42\linewidth}X@{}}
\toprule
Setting & Value\\
\midrule
Reference roster & 16,059 motions\\
Parallelization & 4 GPUs, 4,096 environments per GPU\\
Control / physics & 50\,Hz control, 4 physics substeps per action\\
PPO rollout / minibatch / epochs & 32 steps / 16,384 samples / 6 epochs\\
Actor / critic & Separate ReLU MLPs, $1024\rightarrow1024\rightarrow512$\\
Learning rate / discount / GAE & $2\!\times\!10^{-5}$ / 0.99 / 0.95\\
PPO clip / critic / action-bound coefficients & 0.2 / 5 / 10\\
\bottomrule
\end{tabularx}
\end{table}
Per-object specialists use the same PPO configuration as the generalist in Table~\ref{tab:app_tracker_configuration}, but each trains on the references compatible with its object.

\noindent\textbf{Evaluation.}
Evaluation starts at frame zero from the reference's prescribed initial state, runs to the reference endpoint, and freezes the normalization statistics. Sec.~\ref{app:metrics_tracking} defines the success rate and tracking errors computed from these rollouts.

\FloatBarrier
\section{Self-evolving motion imitation}
\label{app:augmentation}
Augmentation edits a verified reference in interaction coordinates, and the loop keeps only edits that pass physical execution (Sec.~\ref{sec:loop}). Each edit type has its own coordinates, its own way of blending the edit in over time, and its own set of quantities held fixed, while building, execution, and storage are shared by all edits.

\subsection{Edit space and invariants}
\label{app:edit_space}
Edits broaden how each task is physically realized, while its manipulation semantics remain those of the source capture. The edit $\boldsymbol{z}=(\boldsymbol{z}^{\mathrm{task}},\boldsymbol{z}^{\mathrm{body}})$ of Eq.~\ref{eq:interaction_operator} combines spatial coordinates, the object translation and yaw in the seed's interaction frame, with body coordinates that change the pelvis, the stance and toe angle, or the elbow posture. Every edited reference keeps the invariant set $\mathcal I$ of its parent: the interaction type, the intended contacts, the object, and the task outcome. Body edits also keep the object path, palm targets, and relevant support constraints of the parent, and a zero edit reproduces the parent. Each edit is blended in gradually over the interaction through a weight $\alpha_t$, after which the edited reference recomputes its forward kinematics and velocities and its body-object interaction features from the edited geometry (Figure~\ref{fig:bio}).

\subsection{Object edits}
\label{app:verified_augmentation}
Object edits move where the interaction takes place, and the whole body must respond. They act in the seed's interaction frame, whose origin is where a noise-free execution of the seed releases the object, and every edit derived from that seed uses the same frame. With $g_t$ the intended hand contact inherited from the parent, the edit is blended in over the contact phase:
\begin{equation}
 \alpha_t=\begin{cases}
  \displaystyle\frac{\sum_{u\le t}g_u}{\sum_u g_u},&\sum_u g_u>0,\\
  0,&\text{otherwise},
 \end{cases}
 \qquad x'_o(t)=x_o^{\mathrm{parent}}(t)+\alpha_tH_{\mathrm{rel}}\delta,
 \label{eq:app_deformation_ramp}
\end{equation}
where $x_o^{\mathrm{parent}}(t)$ is the object position of the verified parent being edited, $\delta$ is the spatial edit $\boldsymbol z^{\mathrm{task}}$ in the release frame, and $H_{\mathrm{rel}}$ maps it to the world frame. The weight is zero before first contact and one after release, so the edit leaves the approach untouched and accumulates during contact. Moving the goal preserves the type of outcome, such as placement on a support, rather than the exact terminal pose.

\subsection{Body edits}
\label{app:biomechanical}
Body edits vary the robot's posture around a fixed task, and each of their three types keeps its own set of quantities fixed. Pelvis edits change the pelvis and torso configuration, namely crouch, translation, and orientation, and keep the object path, palms, ankle targets, and contact intent. Stance edits change foot separation and toe orientation and keep the root, arms, fingers, object path, and contact intent. Arm edits use the extra freedom of each seven-DoF arm and keep the palms, object, root, legs, fingers, and contact intent. Each edit is blended in and out smoothly around the interaction before the validity checks of Sec.~\ref{app:loop_promotion}.

\noindent\textbf{Fixed-palm arm edits.}
Arm edits move each arm in the nullspace of its palm pose constraint, so the palm stays on its source target. With palm Jacobian $J_h(q_t)\in\mathbb R^{6\times7}$,
\begin{equation}
 J_h(q_t)n_t=0,
 \qquad q_t^{\mathrm{trial}}=q_t+\alpha_t\gamma s_h n_t,
 \label{eq:bio_nullspace}
\end{equation}
where $n_t$ is a nullspace direction, consecutive $n_t$ are sign-aligned, $s_h\in\{-1,1\}$ selects the arm-specific direction, $\gamma$ sets the edit amplitude, and $\alpha_t$ is the edit weight. The constructed pose is projected back to the original palm target before rollout.

\subsection{Building, executing, and keeping edits}
\label{app:loop_promotion}
Object and body edits share the acceptance principle of Sec.~\ref{sec:augmentation}, but differ in when the tracker learns them (Table~\ref{tab:acceptance}). Before physics, a candidate is checked only for basic validity. Object candidates, with near-duplicates removed, then join fine-tuning before they are judged. Body candidates are judged with the current tracker and join training in later rounds once verified. Each candidate is executed with three rollouts under different random seeds.

\begin{table}[ht]
\centering
\small
\caption{\textbf{Acceptance rules for the two edit types.} \checkmark: checked; --: not checked.}
\label{tab:acceptance}
\setlength{\tabcolsep}{5pt}
\begin{tabular}{@{}lcc@{}}
\toprule
& Object edits & Body edits \\
\midrule
Checked before physics & finite values & finite, object unchanged \\
Tracker used & fine-tuned first & current \\
Rollouts needed to pass & 2 of 3 & 1 of 3 \\
\midrule
Completes the reference & \checkmark & \checkmark \\
Terminal hand contact & \checkmark & \checkmark{} (held tasks) \\
No sustained fall & \checkmark & \checkmark \\
Smooth motion & \checkmark & \checkmark \\
Object placed and at rest & -- & \checkmark{} (placed tasks) \\
Object path matches its source & -- & \checkmark\\
\bottomrule
\end{tabular}
\end{table}

\noindent\textbf{Criteria.}
Terminal hand contact holds when the simulated hand contacts match the reference contact labels in six of the final ten frames. A sustained fall is a collapsed or strongly misoriented pose that lasts ten consecutive frames or six of the final ten. Smoothness compares the high-frequency motion of the hands and body, above 6\,Hz, with that of the original source and of the other candidates, and rejects clear outliers. A placed object must rest stably on its support in six of the final ten frames. These checks preserve the type of outcome rather than the exact target pose.

\subsection{Growth over rounds}
\label{app:loop_recursion}
\begin{figure}
    \centering
    \includegraphics[width=\linewidth]{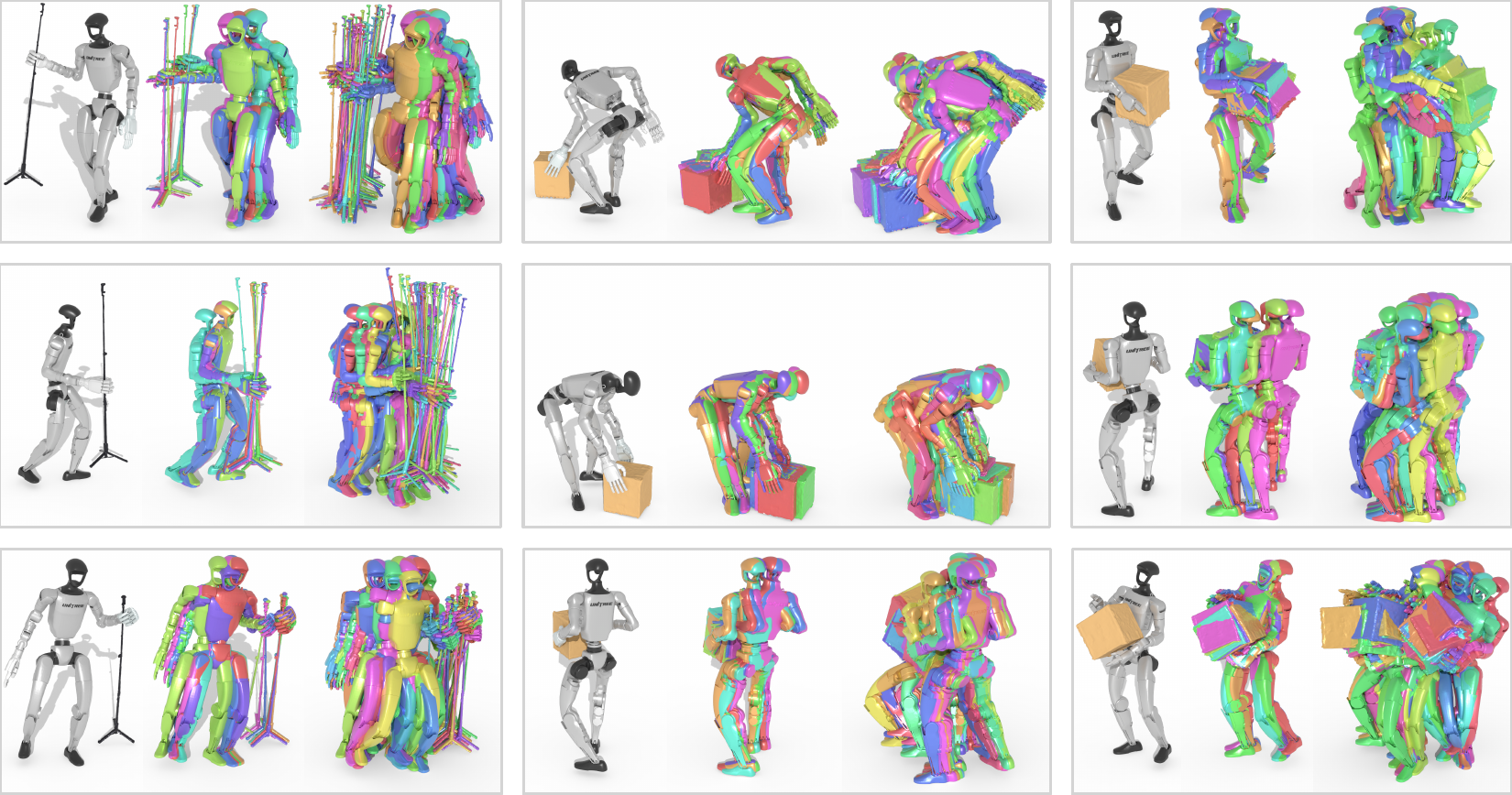}
    \caption{\textbf{More executable variants.} As in Figure~\ref{fig:bio}, for nine further captures: each group shows the original reference followed by the verified augmentations accumulated by round 1 and by round 5.}
    \label{fig:bio_more}
\end{figure}

New edits are proposed only around verified parents, so each round starts where the previous one succeeded. For object edits, each task selects up to 16 parents among the successful edits that reached farthest from their seed, favoring distinct sources and reliable execution, and gives each 16 edits that move further outward. For body edits, each source's successful rollouts are described by their achieved posture, and the most extreme successful postures receive the outward edits of Sec.~\ref{app:biomechanical}. Figure~\ref{fig:bio_more} shows nine further captures.

\section{Evaluation metrics}
\label{app:metrics}
This section defines each metric of Sec.~\ref{sec:eval} once, grouped in the order of the experiments: data coverage (Figure~\ref{fig:data_overview}), retargeted references (Sec.~\ref{sec:retarget_exp}), tracking (Sec.~\ref{sec:tracking_exp}), and the augmentation loop (Sec.~\ref{sec:loop_exp}). All compared methods share the same features, thresholds, and frame rates.

\subsection{Data coverage}
\label{app:metrics_coverage}
The panels of Figure~\ref{fig:pipeline} and Figure~\ref{fig:data_overview} are computed from the following features.

\textbf{Human motion.} Each sequence is described by 21 non-finger joint positions relative to the pelvis, normalized by skeletal height, plus root displacement, height, and heading. Each channel is resampled to 128 frames and truncated to its first 16 DCT coefficients (1,088 dimensions), and PCA-64 then UMAP-2 gives one point per sequence (16,059). The feature uses no object information.

\textbf{Object geometry.} Each mesh is sampled with 10,000 surface points, centered, scaled to the unit sphere, and oriented by its first source pose, then embedded with a frozen OpenShape PointBERT encoder; embeddings are averaged over four yaw rotations. PCA and UMAP give one point per distinct geometry (155). Unit-sphere scaling removes physical size.

\textbf{Human-object interaction.} The feature concatenates the object position relative to the pelvis, the wrist positions and surface distances relative to the object, the wrist separation, and the object height, translation, and rotation, encoded with the same DCT scheme. Each block is normalized separately before PCA-64 and UMAP-2.

\textbf{External coverage.} Probes from GRAIL, MOVIN, and ELMO are matched to the source library in a 73-dimensional body-motion space: PCA-64 of the pelvis-relative joint DCT features plus nine per-limb repetition descriptors. A probe is covered when its nearest source sequence lies within the 75th percentile of the source leave-one-out nearest-neighbor distance, and coverage averages the three probe sets equally. We compare the full InterAct, HiPHI, and their union without subsampling.

\textbf{Scaling.} Coverage is computed on nested random subsets of the union and averaged over 20 orderings; shading marks the 5th--95th percentiles.

\subsection{Retargeted references}
\label{app:metrics_retarget}
These metrics score kinematic references in Table~\ref{tab:retarget_vs_weave} and Table~\ref{tab:g1_omomo_quality}. Each method is evaluated on its own robot model at 30\,fps, on the source clips shared by the compared methods, against the human reference scaled by the subject's body scale.

\textbf{Contact preservation.} During human-labeled hand contact, a hand preserves contact when the minimum distance between its wrist, palm, and finger meshes and the object mesh is at most 2\,cm, with interpenetration counted as zero. The per-frame variant requires all contacting hands to be preserved; the per-hand variant counts each hand--frame separately.

\textbf{In-hand slip.} Mean speed of eleven hand keypoints in the object frame over consecutive contact frames.

\textbf{Penetration.} A frame penetrates when any robot collision geometry is deeper than 1\,cm inside the object (a shared 20-part convex decomposition). We report their mean maximum depth. Ground penetration (Gnd.) is the fraction of frames in which any robot collision geometry is more than 1\,cm below the floor.

\textbf{Foot sliding.} A foot is in stance when the human toe moves horizontally at most 0.15\,m/s, and slides when the robot toe then moves faster than 0.15\,m/s. We report sliding frames over stance frames. Sliding-frame speed is the mean robot toe speed over sliding frames.

\textbf{Fidelity.} MPJPE is the pelvis-relative error of twelve landmarks (hips, knees, ankles, shoulders, elbows, wrists); end-effector error restricts it to the wrists and ankles.

\textbf{Smoothness.} Absolute second finite difference of joint angles times the squared frame rate, averaged over frames and joints. Body (29) and finger (24) joints are reported separately, and the all-joint value averages over all 53. P95 is the 95th percentile of the same quantity over all frames and body joints of a clip, averaged over clips.

\textbf{Joint limits.} Fraction of frames with any joint outside its own range ($10^{-4}$\,rad tolerance). Joint-frame violations count violating joint--frame pairs instead of frames.

\begin{figure}
    \centering
    \includegraphics[width=\linewidth]{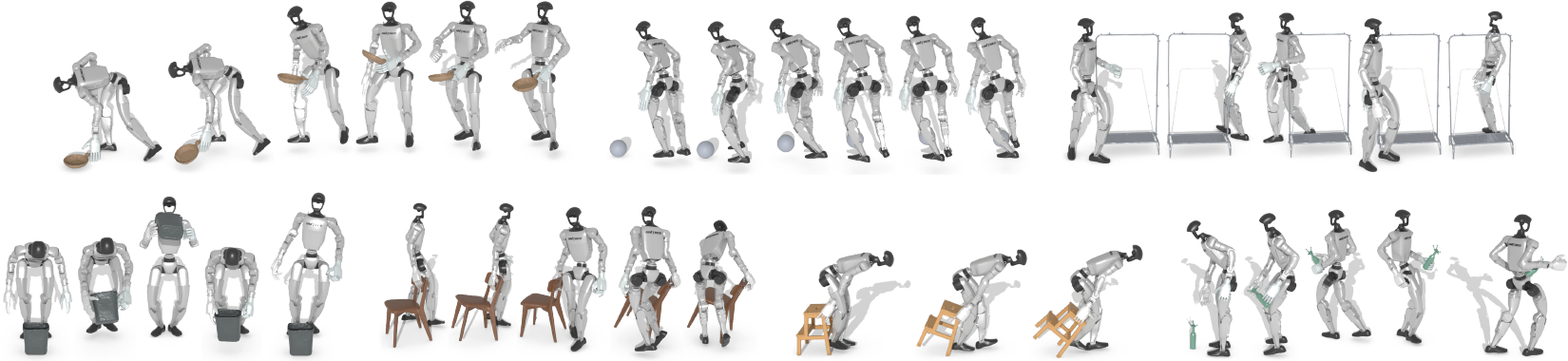}
    \caption{\textbf{More retargeted interactions} on G1 with Inspire hands.}
    \label{fig:retarget_more_objects}
\end{figure}

\begin{figure}
    \centering
    \includegraphics[width=\linewidth]{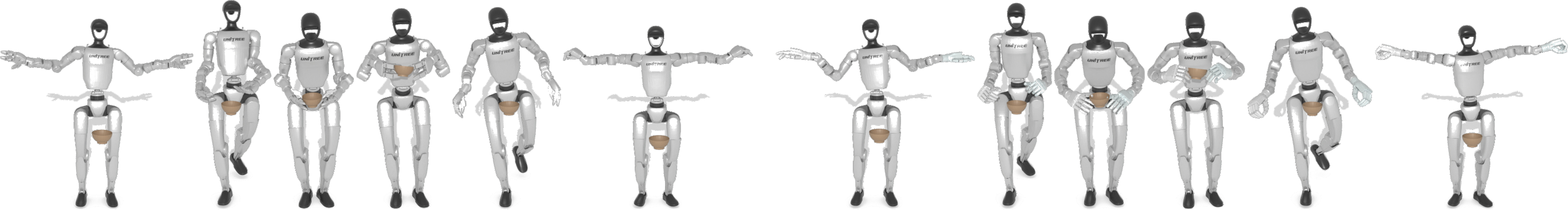}
    \vspace{-1em}
    \caption{\textbf{Retargeting with dexterous hands.} A bowl-holding interaction retargeted to G1 with Dex3 (left) and Inspire (right) hands.}
    \vspace{-1em}
    \label{fig:retarget_dexterous}
\end{figure}

\begin{figure}
    \centering
    \includegraphics[width=\linewidth]{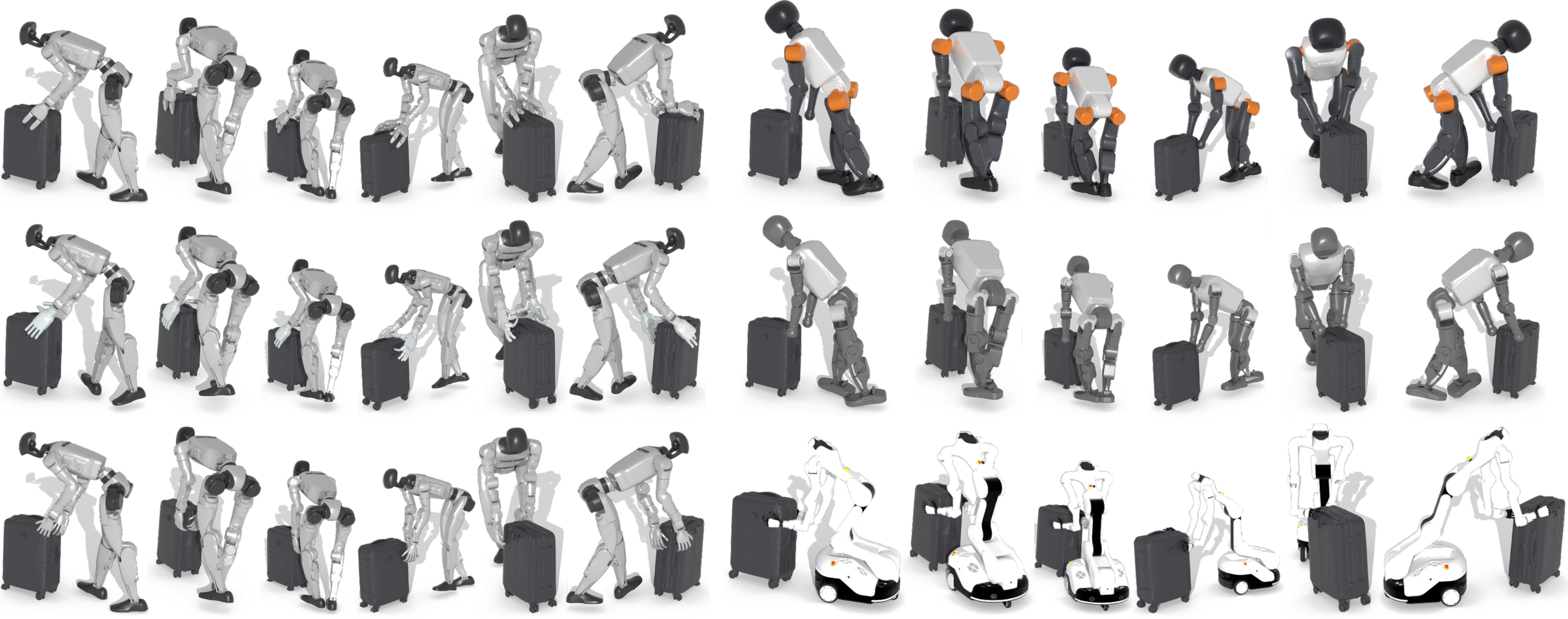}
    \caption{\textbf{Retargeting across embodiments.} One suitcase interaction retargeted to G1 with Dex3, Inspire, and non-dexterous hands (left, top to bottom) and to Booster K1, Booster T1, and Dexmate Vega (right, top to bottom).}
    \label{fig:retarget_embodiments}
\end{figure}

\subsection{Tracking}
\label{app:metrics_tracking}
These metrics score the trackers in Table~\ref{tab:tracking}. Each benchmark clip is executed once under the evaluation protocol of Sec.~\ref{app:verified_protocol}. The benchmark groups its 15 objects into three categories. Bimanual: OMOMO largebox and smallbox, BEHAVE boxlarge and suitcase. Sitting: BEHAVE stool, IMHD chair, and NeuralDome smallsofa. Grasping: OMOMO tripod and largetable, IMHD dumbbell, InterCap cup, racket, and umbrella, GRAB mug, and NeuralDome table.

\textbf{Success rate.} A rollout succeeds when it reaches the last reference frame without triggering a termination condition of Sec.~\ref{app:verified_protocol}. The success rate is the fraction of successful clips.

\textbf{Tracking errors.} Errors are computed on successful clips by comparing each simulated frame with the reference frame of the same index, excluding the terminal hold frames. MPJPE is the mean world-frame position error of the 30 body links shared by all three G1 hand configurations (pelvis, legs, waist, torso, and arms). Object position error is the error of the object origin, and object rotation error is the geodesic angle between the simulated and reference orientations. Per-frame values are averaged within a clip and then across clips.

\subsection{Self-evolution}
\label{app:utility}
These metrics score self-evolution in Tables~\ref{tab:exec_augmented}, \ref{tab:quality_rounds}, and~\ref{tab:frozen_vs_finetuned}. Object and body edits are pooled throughout.

\textbf{Growth.} The number of distinct verified references after a given round, object and body together, divided by the number of original references of the setting.

\textbf{Motion quality.} Quality is computed on the executed trajectory of each original reference and of each reference newly accepted in a round, at 50\,fps and without the 20 terminal-hold frames, and each motion counts equally. Body joint acceleration is the mean absolute second finite difference of the 29 body-joint angles times the squared frame rate. Foot sliding is the fraction of adjacent frames with foot contact in which the ankle link moves horizontally faster than 0.15\,m/s; unlike Sec.~\ref{app:metrics_retarget}, contact is taken from the simulation rather than from the human motion. Hand jitter applies a fourth-order Butterworth high-pass filter above 6\,Hz to pelvis-relative hand positions and reports the 75th percentile of the root-mean-square over 50-frame windows.

\textbf{Success before and after self-evolution.} Each accepted reference and each original reference is replayed with the tracker before round 1 and after round 5, with three rollouts under fixed random seeds and deterministic actions. A reference counts as executed when at least one rollout reaches its last frame without triggering a termination condition of Sec.~\ref{app:verified_protocol}, and the success rate is the fraction of references executed.

\textbf{Yield.} In the frozen-tracker ablation, the yield is the fraction of a round's rollouts that pass the acceptance checks.

\section{Additional results}
\label{app:exp}
Additional tables and figures extend the retargeting, tracking, and loop comparisons of Sec.~\ref{sec:experiments}, and a frozen-tracker ablation isolates the role of fine-tuning in the loop.

\begin{figure}
    \centering
    \includegraphics[width=\linewidth]{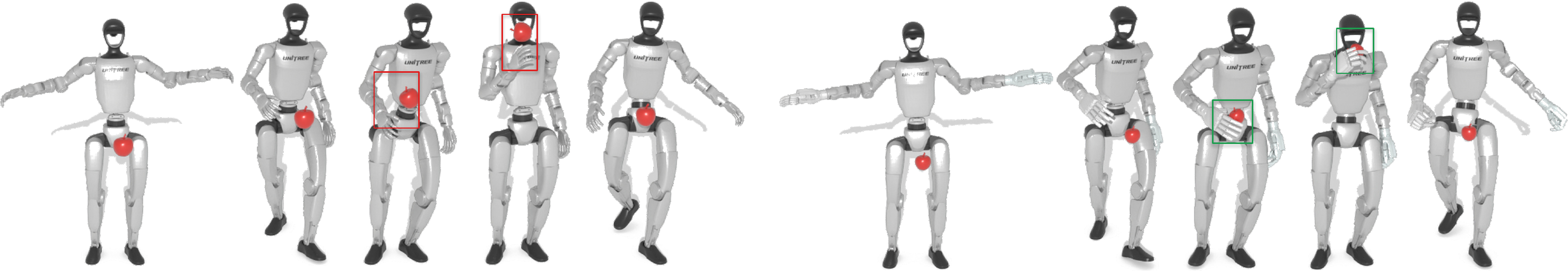}
    \caption{\textbf{Retargeting comparison with OmniRetarget.} OmniRetarget to G1 with rigid hands (left) and our retargeter for G1 with Inspire hands (right), at corresponding interaction phases. The dexterous hand keeps the grasp and the hand-object relationship of the demonstration.}
    \label{fig:retarget_omni}
\end{figure}

\begin{figure}
    \centering
    \includegraphics[width=\linewidth]{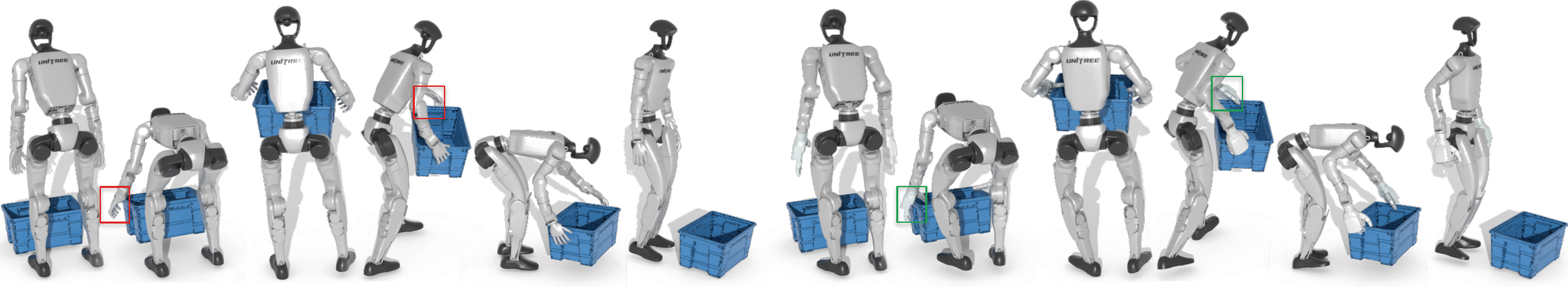}
    \caption{\textbf{Retargeting comparison with UMR.} UMR to G1 (left) and our retargeter to G1 with Inspire hands (right), at corresponding interaction phases.}
    \label{fig:retarget_umr}
\end{figure}

\subsection{Retargeting comparison on G1 with non-dexterous hands}
\label{app:rubber_hand}
All four retargeters are compared on G1 with non-dexterous hands, using the 537 OMOMO clips they share (largebox, smallbox, plasticbox, and suitcase) and the metrics of Sec.~\ref{app:metrics_retarget}. Our entries use the final cleaned references for every object. Contact preservation uses the supplied left/right human-hand contact labels, which are positive in 479 clips. ULTRA produces several variants per source sequence, which we average before averaging all metrics over source sequences. ULTRA's variants have no matching human reference, so their foot-sliding and fidelity metrics are unavailable.

\begin{table}[!htbp]
  \caption{\textbf{Retargeting on G1 with non-dexterous hands}, over the 537 OMOMO clips shared by all four methods. A dash denotes no qualifying event and N/A an unavailable metric; bold marks the unique best.}
  \label{tab:g1_omomo_quality}
  \centering
  \small
  \begin{tabular}{lcccc}
  \toprule
  Metric & OmniRetarget & UMR & ULTRA & Ours \\
  \midrule
  \multicolumn{5}{l}{\textit{Contact preservation}} \\
  \quad Hand contact preservation (\%) $\uparrow$
      & 79.09 & \textbf{95.13} & 79.06 & 83.45 \\
  \addlinespace
  \multicolumn{5}{l}{\textit{Penetration ($>1$\,cm)}} \\
  \quad Robot--object depth (cm) $\downarrow$
      & \textbf{1.27} & 1.60 & 1.49 & 1.97 \\
  \addlinespace
  \multicolumn{5}{l}{\textit{Foot sliding (0.15\,m/s)}} \\
  \quad Sliding frames (\%) $\downarrow$
      & \textbf{0.00} & 7.02 & N/A & 7.33 \\
  \quad Sliding-frame speed (m/s) $\downarrow$
      & -- & 0.22 & N/A & 0.26 \\
  \addlinespace
  \multicolumn{5}{l}{\textit{Fidelity to human}} \\
  \quad MPJPE (cm) $\downarrow$
      & \textbf{5.55} & 6.70 & N/A & 5.62 \\
  \quad End-effector error (cm) $\downarrow$
      & 6.36 & 7.92 & N/A & \textbf{5.86} \\
  \addlinespace
  \multicolumn{5}{l}{\textit{Smoothness}} \\
  \quad Body joint acc. (rad/s$^2$) $\downarrow$
      & 6.65 & 6.11 & 10.54 & \textbf{2.54} \\
  \quad Body joint acc., P95 (rad/s$^2$) $\downarrow$
      & 23.19 & 21.97 & 37.29 & \textbf{8.96} \\
  \addlinespace
  \multicolumn{5}{l}{\textit{Joint limits}} \\
  \quad Body violation frames (\%) $\downarrow$
      & 0.00 & 0.00 & 1.04 & 0.00 \\
  \quad Body joint-frame violations (\%) $\downarrow$
      & 0.00 & 0.00 & 0.04 & 0.00 \\
  \bottomrule
  \end{tabular}
\end{table}

\noindent\textbf{With rigid hands, each method favors a different trade-off.}
Table~\ref{tab:g1_omomo_quality} shows no single winner between fidelity, geometric constraints, and smoothness. Our references are by far the smoothest and the most accurate at the end effectors, OmniRetarget penetrates the object least, and UMR preserves the most hand contact. Rigid hands cannot close around an object, so contact and penetration trade off directly. This is the gap our dexterous retargeting addresses: with articulated fingers, our references keep the grasp and hand--object relationship of the demonstration (Figures~\ref{fig:retarget_omni} and~\ref{fig:retarget_umr}).

\begin{figure}
    \centering
    \includegraphics[width=\linewidth]{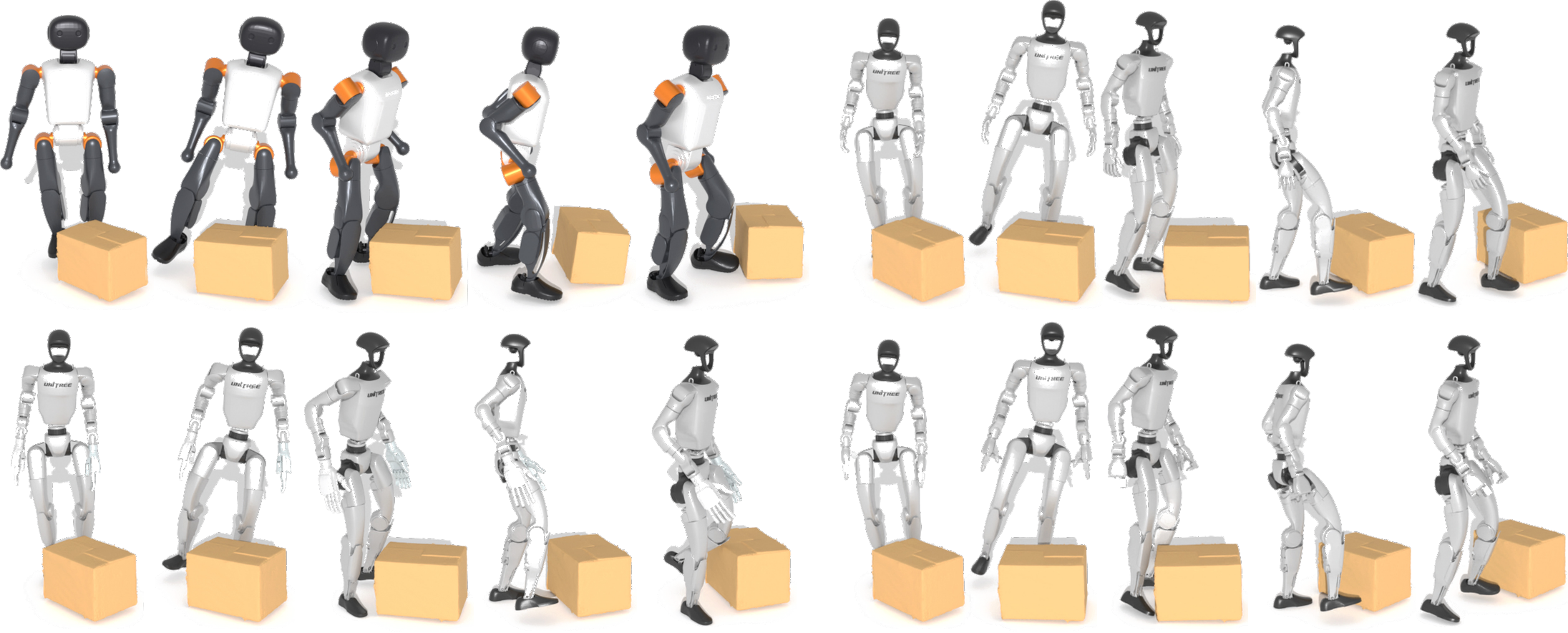}
    \caption{\textbf{Physics-based tracking across embodiments.} A box-lifting reference executed by specialist trackers on Booster K1 (top left), G1 with non-dexterous hands (top right), G1 with Inspire hands (bottom left), and G1 with Dex3 hands (bottom right).}
    \label{fig:tracking_embodiments}
\end{figure}

\subsection{Ablation on learning in the loop}
\label{sec:ablation_finetune}
\noindent\textbf{More sampling cannot replace learning.}
A fixed tracker can only filter candidates it already executes, and after the first round it has nothing left to filter. Because the base tracker fails on many accepted references (Table~\ref{tab:exec_augmented}), a simpler alternative to fine-tuning would keep the tracker fixed, sample more rollouts, and keep those that pass. Table~\ref{tab:frozen_vs_finetuned} tests this alternative in a separate run on a subset of the bimanual seeds. In the first round, both trackers execute the same candidates with the same checkpoint, so their yields are similar. From the second round on, the frozen tracker accepts almost nothing, because the new candidates start from first-round successes that already lie beyond its reach, and its growth stalls. The fine-tuned tracker keeps a high yield in rounds 2 and 3, so its verified references keep growing.

\begin{table}[t]
\caption{\textbf{Frozen versus fine-tuned tracker} in a separate run on a subset of the bimanual seeds with translation edits, whose counts are not directly comparable with Table~\ref{tab:exec_augmented}. Yield and growth follow Sec.~\ref{app:utility}. The frozen run stops after round 3.}
\label{tab:frozen_vs_finetuned}
\centering
\small
\setlength{\tabcolsep}{0pt}
\begin{tabular*}{0.7\linewidth}{@{\extracolsep{\fill}}ccccc@{}}
\toprule
& \multicolumn{2}{c}{Yield (\%)} & \multicolumn{2}{c}{Growth (\texttimes)} \\
\cmidrule(lr){2-3}\cmidrule(l){4-5}
Round & Frozen & Tuned & Frozen & Tuned \\
\midrule
1 & 29.4 & 25.3 & 163.0 & 188.5 \\
2 & 0.4 & 72.9 & 170.5 & 426.0 \\
3 & 0.4 & 68.9 & 184.9 & 573.0 \\
\bottomrule
\end{tabular*}
\end{table}

\subsection{Self-evolution by round}
\label{app:quality_rounds}
\noindent\textbf{The reference set grows every round, while quality measures rise moderately.}
Table~\ref{tab:quality_rounds} extends Table~\ref{tab:exec_augmented} to every round. Growth is measured on the cumulative set of verified references, and quality on the references newly accepted in each round. The verified set grows in every round in all three settings. All three quality measures rise in later rounds, most clearly foot sliding; later rounds, however, accept edits farther from the captures and a different mix of object and body edits, so the trend reflects harder motions as well as execution quality. Success rates are measured only before and after self-evolution and appear in Table~\ref{tab:exec_augmented}.

\begin{figure}
    \centering
    \includegraphics[width=\linewidth]{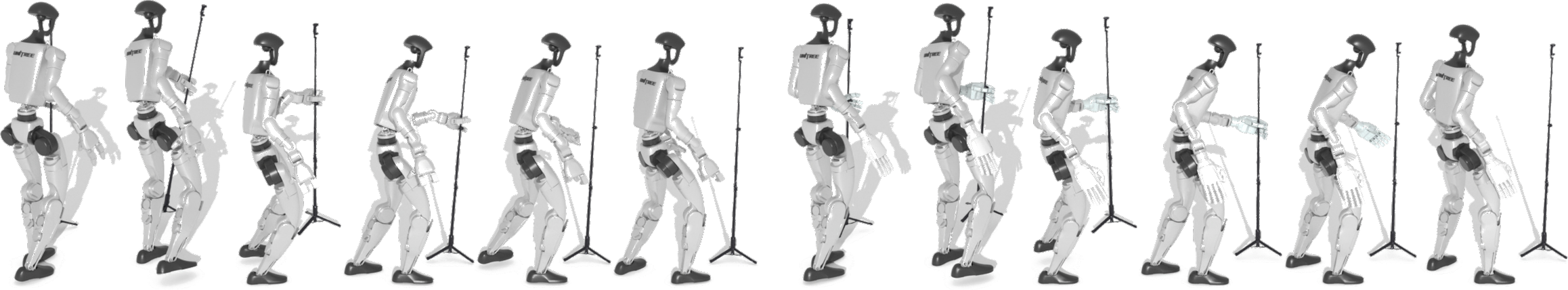}
    \caption{\textbf{Physics-based tracking with dexterous hands.} A tripod grasp executed on G1 with Dex3 (left) and Inspire (right) hands.}
    \label{fig:tracking_dexterous}
\end{figure}

\begin{table}
\caption{\textbf{Self-evolution by round}, extending Table~\ref{tab:exec_augmented}. Growth (\texttimes): cumulative verified references relative to the original references. Quality of the original references (Orig.) and of the references newly accepted in each round: body joint acceleration (Acc., rad/s$^2$), foot sliding (Foot, \%), and hand jitter (Hand, mm).}
\label{tab:quality_rounds}
\centering
\footnotesize
\setlength{\tabcolsep}{0pt}
\begin{tabular*}{\linewidth}{@{\extracolsep{\fill}}lcccccccccccc@{}}
\toprule
& \multicolumn{4}{c}{Inspire, bimanual} & \multicolumn{4}{c}{Dex3, bimanual} & \multicolumn{4}{c}{Inspire, grasping} \\
\cmidrule(lr){2-5}\cmidrule(lr){6-9}\cmidrule(l){10-13}
Round & Growth & Acc. & Foot & Hand & Growth & Acc. & Foot & Hand & Growth & Acc. & Foot & Hand \\
\midrule
Orig. & 1.0 & 11.4 & 1.18 & 1.00 & 1.0 & 13.9 & 1.14 & 1.12 & 1.0 & 13.6 & 0.70 & 0.96 \\
1 & 21.8 & 11.8 & 1.39 & 1.07 & 21.3 & 15.6 & 1.59 & 1.25 & 19.6 & 15.3 & 0.99 & 1.07 \\
2 & 34.8 & 11.3 & 1.39 & 1.09 & 31.6 & 14.8 & 1.77 & 1.23 & 33.3 & 14.4 & 1.42 & 1.09 \\
3 & 80.4 & 12.1 & 1.63 & 1.16 & 73.3 & 14.9 & 1.87 & 1.24 & 77.8 & 14.6 & 1.78 & 1.08 \\
4 & 118.1 & 12.9 & 2.14 & 1.22 & 110.8 & 16.4 & 2.41 & 1.33 & 112.8 & 15.7 & 2.42 & 1.16 \\
5 & 150.5 & 13.2 & 2.37 & 1.24 & 142.0 & 17.4 & 2.99 & 1.38 & 146.4 & 16.4 & 2.87 & 1.26 \\
\bottomrule
\end{tabular*}
\end{table}

\subsection{Body-motion range of augmented references}
\label{app:body_range}
\begin{figure}
\centering
\includegraphics[width=0.55\linewidth]{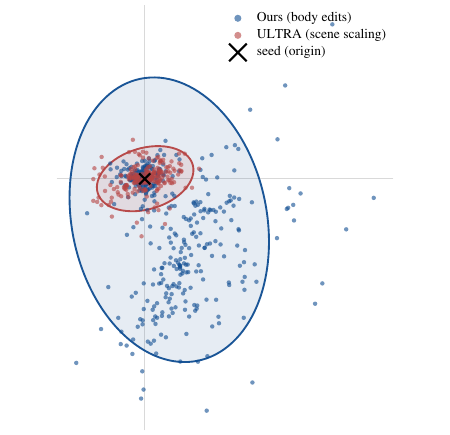}
\caption{\textbf{Body-motion variation around shared seeds.} Each reference is embedded by the DCT coefficients of its body joint angles, taken relative to its seed and projected with PCA, so all seeds sit at the origin. Ellipses enclose 95\% of a Gaussian fit.}

\label{fig:body_motion_ultra}
\end{figure}

\noindent\textbf{Body edits move references farther from their seed than scene scaling does.}
Figure~\ref{fig:body_motion_ultra} compares how far augmented references depart from their seed in body motion, on the bimanual seeds shared by both methods. Our body edits spread over a region several times wider than ULTRA's scene-scaling variants, and mostly along directions that scaling does not vary.

\section{Hardware and real-world deployment}
\label{app:hardware}

\noindent\textbf{Deployment for Unitree G1.} Experiments are performed on a Unitree G1 humanoid with 29 DOFs (six per leg, three at the waist, and seven per arm), running robot firmware version 1.5.4. The policy is executed on a laptop with an NVIDIA RTX 5090 Laptop GPU and communicates over Ethernet through Unitree's SDK. We use SONIC~\citep{sonic} as the low-level whole-body controller: it takes the reference motion at 50\,Hz, which is tracked by the onboard joint-level PD controllers at 500\,Hz, using SONIC's default per-joint PD gains. On top of these, stiffness and damping scales of 1.5 are applied to the two ankle pitch motors; a constant 1 degree pitch offset is further added to the measured pelvis IMU orientation to account for the calibration bias of the robot's own IMU.

\noindent\textbf{Deployment for Unitree G1 with Inspire dexterous hands.} For tasks that require dexterous manipulation, we equip Unitree G1 with two Inspire RH56DFTP dexterous hands, each with six actuated DOFs, plus six mimic joints. The humanoid body is controlled by SONIC as described above, while the hands directly replay the finger trajectory of the rollout motion. For the synchronization, the body deployment publishes its current playback frame, and a separate process sends the hand pose for that frame to the robot at about 12\,Hz. Notably, we initiate the thumb motion first, followed by the motion of the other four fingers.

\noindent\textbf{Deployment for Booster K1.} Experiments are performed on a Booster K1 humanoid with 22 DOFs (six per leg, four per arm, and two at the head). For policy training, we build on Booster Train (\url{https://github.com/BoosterRobotics/booster_train}), extending it with object-interaction tracking and targeted domain randomization. We use an asymmetric actor-critic formulation, with privileged object and contact information provided only to the critic during training. Robot and object dynamics are randomized, including ground friction, object mass, center of mass, object placement, actuator properties, and external perturbations, while preserving close tracking of the reference body and object motion. For real-world execution, policies are deployed using Booster Deploy (\url{https://github.com/BoosterRobotics/booster_deploy}).

\noindent\textbf{Deployment for Dexmate Vega.} Experiments are performed on a Dexmate Vega-1 Pro, a bimanual mobile manipulator
with 23 DOFs (three at the torso, seven per arm, three at the head, and three DOFs at the omnidirectional wheeled base). Since the wheeled base is stable and agile, no whole-body controller from RL is needed. Retargeted upper-body joint trajectories are sent directly to the onboard joint controllers through Dexmate's dexcontrol SDK (v0.5.0). Notably, due to the torso joint speed limit, all trajectories are uniformly time-scaled by $0.3\times$ speed during deployment. Base motion is converted to wheel commands by solving the chassis inverse kinematics using trajectory optimization, formulated in CasADi~\citep{Andersson2019} and solved with IPOPT~\citep{wachter2006implementation}. The solver penalizes position, yaw, terminal-pose tracking errors, and regularizes body acceleration, steering rate and acceleration, and wheel acceleration. All commands are streamed at 100\,Hz.

\section{Limitations} The loop varies how an interaction is realized, so it densifies coverage around each demonstration without creating new task semantics. Iterative selection may also favor easier objects or compound reference errors.

\end{document}